\documentclass[sigconf]{acmart}

\copyrightyear{2020} 
\acmYear{2020} 
\setcopyright{acmcopyright}\acmConference[CIKM '20]{Proceedings of the 29th ACM International Conference on Information and Knowledge Management}{October 19--23, 2020}{Virtual Event, Ireland}
\acmBooktitle{Proceedings of the 29th ACM International Conference on Information and Knowledge Management (CIKM '20), October 19--23, 2020, Virtual Event, Ireland}
\acmPrice{15.00}
\acmDOI{10.1145/3340531.3412733}
\acmISBN{978-1-4503-6859-9/20/10}

\usepackage{subcaption}
\usepackage{multirow}
\usepackage{makecell}
\AtBeginDocument{%
  \providecommand\BibTeX{{%
    \normalfont B\kern-0.5em{\scshape i\kern-0.25em b}\kern-0.8em\TeX}}}

\newcommand{\haotian}[1]{\textcolor{black}{#1}}
\mathchardef\mhyphen="2D

\begin{document}
\fancyhead{}
\title{Peer-inspired Student Performance Prediction in Interactive Online Question Pools with Graph Neural Network}

\author{Haotian Li, Huan Wei, Yong Wang, Yangqiu Song, Huamin Qu}
\affiliation{%
  \institution{Department of Computer Science and Engineering, HKUST, Hong Kong SAR, China}
}
\email{{haotian.li, hweiad, ywangct}@connect.ust.hk;{yqsong, huamin}@cse.ust.hk}

\begin{abstract}

% 1. The importance of student performance prediction
% 2. Challenges in conducting student performance prediction in interactive online question pools
% 3. Our approach
% 4. Evaluations

Student performance prediction is critical to online education. It can benefit
%facilitate
many downstream tasks on online learning platforms, such as estimating dropout rates, 
% keeping student retention, 
facilitating strategic intervention, and enabling adaptive online learning.
\textit{Interactive online question pools} provide students with interesting interactive questions to practice their knowledge in online education.
However, little research has been done on student performance prediction in interactive online question pools. 
Existing work on student performance prediction targets at online learning platforms with predefined course curriculum and accurate knowledge labels like MOOC platforms,
but they are not able to fully
% can only partially
 model knowledge evolution of students in interactive online question pools.
%Most existing work on student performance prediction targets at online learning platforms with predefined course curriculum like MOOC platforms, and intrinsically relies on the sequential order of learning materials (e.g., videos and questions).
%These prediction methods cannot work for interactive online question pools, as no predefined sequential order exists, and students often need to freely select which question to answer next in interactive online question pools.
% often need to freely select which question to practice next,  
% Different from them,
% no predefined sequential order exists in interactive online question pools,
% % often need to freely select which question to practice next, 
% making it difficult to apply prior 
% student performance prediction algorithms to interactive online question pools.
% Some recent research conducted student performance prediction on interactive online question pools by modeling the question similarity based on mouse interaction features, but it
% is limited to
% questions of the same type and cannot be generalized to other questions.
In this paper, we
propose a novel approach using Graph Neural Networks (GNNs) to achieve better student performance prediction in interactive online question pools. 
Specifically,
we model the relationship between students and questions using student interactions to construct the \textit{student-interaction-question} network and further present a new GNN model, called R$^2$GCN, which intrinsically works for the heterogeneous networks, to achieve generalizable student performance prediction in interactive online question pools.
We evaluate the effectiveness of our approach on a real-world dataset consisting of 104,113 mouse trajectories generated in the problem-solving process of over 4,000 students on 1,631 questions. The experiment results show that our approach can achieve a much higher accuracy of student performance prediction than both traditional machine learning approaches and GNN models.

% Detailed experiments on a real-world dataset demonstrate the effectiveness of our approach compared with both traditional machine learning approaches and GNN models.

\end{abstract}


% \ccsdesc[500]{Applied computing~E-learning}
% \ccsdesc[500]{Applied computing~Interactive learning environments}
% \ccsdesc[300]{Applied computing~Learning management systems}
% % \ccsdesc[500]{Applied computing~Collaborative learning}
% \ccsdesc[500]{Computing methodologies~Neural networks}
% \ccsdesc[300]{Computing methodologies~Semi-supervised learning settings}

\keywords{Student performance prediction, graph neural networks, online question pools}

\maketitle

\section{Introduction}
% New Overall Framework:
% 1. The importance of student performance prediction in online education
% 2. Limitations of existing work on learning modeling and student performance prediction: 1) Traditional methods, 2) DKT, etc.
% 3. challenges in student performance prediction in interactive online question pools. Huan's work
% 4. Our approach
% 5. Evaluation.

%The past few years have seen the rapid growth of online learning. Various online learning platforms, including interactive online question pools and massive open online courses (MOOCs), are becoming increasingly popular.
%Student performance prediction
%% , a well-established task in online education, 
%aims to predict students' future grades in the course assignments, quizzes and final exams by using their historical data~\cite{DBLP:conf/edm/HuR19}.
%% Student performance prediction is a well-established task in online education.
%% Given the historical data of students such as course grades and attempted question lists, the main goal of student performance prediction is to predict students' grades in the course assignments, quizzes and final exams~\cite{DBLP:conf/edm/HuR19}.
%% % The past few years have seen the rapid growth of various online learning platforms such as interactive online question pools and massive open online courses (MOOCs). 
%% Student performance prediction on the online learning platforms

With the rapid growth of online education in the past few years, various online learning platforms (e.g.,  interactive online question pools) have become increasingly popular.
Student performance prediction, which aims to predict students' future grades in the assignments and exams~\cite{DBLP:conf/edm/HuR19}, is of significant importance for online education.
%It has significant importance in online education.
For example, with an accurate student performance prediction,
the online learning platforms can better estimate the course dropout rate and take appropriate measures to further increase the student retention rate~\cite{RenRJ16}. The course instructors can recommend suitable learning materials to different students~\cite{recommend}.
%The course designers and instructors can conduct strategic interventions to help students who may face troubles in the learning process
%% as teachers do in university education
%% ~\cite{strategicIntervention} 
%and further provides students of different learning capabilities and knowledge levels with personalized education by recommending suitable learning materials~\cite{recommend}.
%% enhancing the effectiveness of online learning.
Extensive research has been conducted on student performance prediction on online learning platforms, which mainly consists of static models and sequential models~\cite{DBLP:conf/edm/HuR19}.
%which generally can be categorized into two types~\cite{DBLP:conf/edm/HuR19}: static models and sequential models.
Static models consider the static information of students (e.g., historical scores and learning activities) to predict their future performance~\cite{RenRJ16, jiang14, KennedyCBC15}, where the underlying relationship between different learning materials (e.g., courses, videos, questions) are totally ignored. 
%Sequential models advance static models by further capturing the sequential relationship of the learning materials. The representative methods are the Recurrent Neural Network (RNN) based approaches, such as Deep Knowledge Tracing (DKT)~\cite{DKT} and its variant methods~\cite{yeung2018addressing, Chen0ZP18}. 
%Nevertheless, these RNN-based approaches are mainly applied to MOOC platforms, and intrinsically rely on the sequential order of the course materials defined by the curriculum. 
Sequential models, such as Deep Knowledge Tracing (DKT)~\cite{DKT} and its variant methods~\cite{GKT, Chen0ZP18}, further capture the sequential relationship of the learning materials. However, the sequential models are mainly applied to MOOC platforms and intrinsically rely on accurate labeling of the tested knowledge for each question.

Interactive online question pools, an essential part of online learning, attempts to make it a joyful process for students to practice their knowledge on a collection of interactive questions.
For instance, Math Playground\footnote{\href{https://www.mathplayground.com/}{https://www.mathplayground.com/}}, Learnlex\haotian{\footnote{\href{ https://learnlex.com/}{https://learnlex.com/}}}, and LeetCode\footnote{\href{https://leetcode.com/}{https://leetcode.com/}} enable students to practice their mathematics or programming skills.
%For instance, Math Playground\footnote{\href{https://www.mathplayground.com/}{https://www.mathplayground.com/}} and Learnlex\footnote{\href{ https://mad.learnlex.com/login}{https://mad.learnlex.com/login}} give students opportunities to practice their mathematical skills. LeetCode\footnote{\href{https://leetcode.com/}{https://leetcode.com/}} and TopCoder\footnote{\href{https://www.topcoder.com/}{https://www.topcoder.com/}} enable students to improve their coding capabilities.
However, interactive online question pools are different from MOOC platforms and there is \textit{no predefined sequential order} for the learning materials (i.e., questions). Students often need to 
%explore the questions by themselves and 
freely choose which question to answer next. 
It makes the prior sequential models only able to partially model the student knowledge evolution.
Also, an accurate label of the tested knowledge for each question is also not necessarily available in interactive online question pools~\cite{Wei2020lak}. 
Therefore, we are motivated by the crucial research question: \textit{how can we achieve effective student performance prediction in interactive online question pools?}

In this paper, we propose a novel GNN-based approach
% based on GNNs 
% for student performance prediction 
to model student knowledge evolution and predict student performance
 in interactive online question pools.
%It is intrinsically able to avoid dependency on predefined question order.
Specifically, 
we first build a heterogeneous large graph, consisting of questions, students, and the interactions between them, to extensively model the complex relationship among different students and questions. This is inspired by prior studies~\cite{kassarnig2017class,kassarnig2018academic} and they have shown that the academic performance of a student is correlated with the performances of other students (i.e., \textit{peers}), especially those students with similar learning behavioral patterns.
%Specifically,
%we extensively model the underlying relationship among questions and students to achieve better performance prediction accuracy in interactive online question pools.
%Prior research on academic performance~\cite{kassarnig2017class,kassarnig2018academic} has shown that the academic performance of a student is correlated with the performances of other students (i.e., \textit{peers}), especially those students with similar learning behavioral patterns.
%Thus, we first build a heterogeneous large graph consisting of questions, students and the interactions between them to model the complex relationship among different students and questions, and further
%formalize student performance prediction as a semi-supervised node classification problem on this graph.
%% a heterogeneous large graph consisting of questions, students and the interactions between them, 
%The classification results are the student score level (4 score levels in our experiment) on each question. We also extend the mouse movement features in prior work~\cite{Wei2020lak} to further consider mouse click interactions in students' problem-solving processes.
Then, we further
formalize student performance prediction as a semi-supervised node classification problem on this heterogeneous graph.
% a heterogeneous large graph consisting of questions, students and the interactions between them, 
The classification results are the student score levels (4 score levels in our experiment) on each question.
%, where the mouse movement features adapted from prior work~\cite{Wei2020lak}  is also considered.
% , which are more widely existing. 
Moreover, we propose a novel GNN model, \textit{Residual Relational Graph Neural Network (R$^2$GCN)}, for student performance prediction in interactive online question pools.
%to predict a student's score level on each question in interactive online question pools. 
Its model architecture is adapted from Relational-GCN (R-GCN)~\cite{Schlichtkrull2017ModelingRD} and further incorporates a residual connection to different convolutional layers and original features.
% into the architecture.
We conduct detailed evaluations of our approach
% and compare it with state-of-the-art student performance prediction methods 
 on a real-world dataset consisting of 104,113 mouse trajectories generated in the problem-solving process of over 4,000 students on 1,631 questions, which are collected by our industry collaborator Trumptech\footnote{\href{https://www.trumptech.com/en}{https://www.trumptech.com/en}} from their K-12 interactive online math question pool \haotian{\textit{Learnlex}}.
%
% consisting of 104,113 mouse trajectories generated in the problem-solving process of over 4000 students on 1631 questions, which are provided by our industry collaborator Trumptech~\footnote{\href{https://www.trumptech.com/en}{https://www.trumptech.com/en}} and collected from their K-12 interactive online math question pool, \textit{LearnLex}.
% 
The results show that our approach outperforms other methods in terms of both accuracy and weighted F1 score.
Detailed insights and observations are also discussed.
In summary, the major contributions are as follows:
% \vspace{-1em}
\begin{itemize}
\item \textbf{Question Formulation.} 
%To achieve a more accurate student performance prediction, we aim to 
%we capture the underlying relationship among questions and students and
% To better capture the underlying relationship among questions and students, we 
We formulate the student performance prediction in interactive online question pools as a semi-supervised node classification problem on a large heterogeneous graph that captures the underlying relationship among questions and students.
% consisting of questions, students and the interactions between them.
New mouse movement features are also introduced to better delineate student-question interactions.
%%%%%%%%%%%%%%%%%%%%%%%%%%%%%%%%%

\item \textbf{Model Architecture.} We propose a new convolutional graph neural network model, R$^2$GCN, to achieve student performance prediction in interactive online question pools, which intrinsically works for heterogeneous networks.

\item \textbf{Detailed Evaluations.} We conduct detailed evaluations of our approach on a real-world dataset. The results demonstrate its capability of achieving a better prediction accuracy than both traditional machine learning models (e.g., Logistic Regression (LR) and Gradient Boosted Decision Tree (GBDT) model), and R-GCN~\cite{Schlichtkrull2017ModelingRD}.

\end{itemize}
\nopagebreak[4]

\section{Related Work} \label{related}
The related work of this paper can be categorized into two groups: Graph Neural Networks and student performance prediction.
% , and mouse interaction feature extraction.
\vspace{-0.5em}
\paragraph{Graph Neural Networks}

Graph Neural Networks (GNNs) are the deep neural networks adapted from the widely-used Convolutional Neural Networks (CNNs) and specifically designed for graphs. They 
% often apply a message passing scheme to modeling the complicated relationship between nodes and 
have shown powerful capability in dealing with complicated relationships in a graph.
% and some representative works are 
Representative methods of GNNs include
Graph Convolutional Network (GCN)~\cite{kipf2017semi}, GraphSAGE~\cite{hamilton2017inductive}, R-GCN~\cite{Schlichtkrull2017ModelingRD}, Message Passing Neural Network (MPNN)~\cite{Gilmer2017NeuralMP}, Gated Graph Neural Network (GGNN)~\cite{Li2016GatedGS}, and Heterogeneous Graph Attention Network (HAN)~\cite{han}. Among them, R-GCN and HAN are specifically designed for heterogeneous graphs. MPNN and GGNN perform graph convolution on graphs with multi-dimensional edge features. However,
little research has been conducted on heterogeneous graphs with multi-dimensional edge features.

% there are few works on heterogeneous graphs with multi-dimensional edge features.

GNNs have been applied in various applications, such as recommender systems~\cite{RecommenderSys}, social networks analysis~\cite{SocialNet}, and molecular property prediction~\cite{Li2016GatedGS}.
Very few studies have been done in the field of online learning and education. A recent study on college education by Hu and Rangwala~\cite{DBLP:conf/edm/HuR19} proposed a GNN-based approach called Attention-based Graph Convolutional Network (AGCN) which utilizes a GCN to learn graph embedding of the network of frequently taken prior courses and then applies the attention mechanism to generate weighted embedding for final predicted grades.
However, their method is limited to the graph with only one type of nodes (i.e., courses) and edges (i.e., the connection of courses taken in continuous 2 semesters), which cannot be applied to student performance prediction in interactive online question pools due to their intrinsically complex relationships among questions and students.

% Nowadays, GNNs have been widely applied in various fields, including social networks analysis~\cite{SocialNet}, and prediction of molecular properties~\cite{Jin2018LearningMG}. Currently, the application of GNN in the area of education is still underexplored. Hu and Rangwala proposed AGCN~\cite{DBLP:conf/edm/HuR19} for students' grade prediction of undergraduate courses, which is the only work in this area. 
% However, the model is not suitable for grade prediction in interactive online question pools. In the graphs of undergraduate courses, there only exists one type of edge representing pre-requisite and one type of node representing courses, which means the model could not be applied to our heterogeneous graphs of students, questions and interactions. To the best of our knowledge, we are the first to utilize a GNN on heterogeneous networks for students’ grade prediction.

\vspace{-0.5em}
\paragraph{Student performance prediction}

Student performance prediction is an important task in educational data mining. For example, it can contribute to recommending learning material~\cite{recommend} and improving student retention rates~\cite{RenRJ16} in online learning platforms.
According to the study by Hu and Rangwala~\cite{DBLP:conf/edm/HuR19}, prior studies on student performance prediction mainly include static models and sequential models.
% Various works on this task could be mainly categorized into 2 classes, static models and sequential models according to Hu and Rangwala's study~\cite{DBLP:conf/edm/HuR19}. 
Static models refer to traditional machine learning models such as GBDT~\cite{Wei2020lak}, Supporting Vector Machine (SVM)~\cite{SVM}, and LR~\cite{LR}, which make predictions on student performances based on the static patterns of student features.
% The representative algorithms of this type include GBDT~\cite{Wei2020lak}, Supporting Vector Machine (SVM)~\cite{SVM} and LR~\cite{LR}.
\haotian{On the contrary, sequential models~\cite{DKT, Chen0ZP18, GKT} are proposed to better capture the temporal evolutions in students' knowledge or the underlying relationship between learning materials}.
% Under this category, 
% For instance,
% Deep Knowledge Tracing (DKT)~\cite{DKT} and its several variants~\cite{yeung2018addressing, Chen0ZP18, GKT} introduce RNNs on a sequence of students' problem-solving records to extract the hidden knowledge and model their knowledge evolution.
% These models often require a fixed order of the learning materials, making them intrinsically rely on the predefined curriculum of the courses. 
% knowledge components in well-designed curriculum. 
% For instance, 
% to track students' mastering of knowledge components,
% Bayesian Knowledge Tracing (BKT) had first been proposed~\cite{BKT0} using given knowledge concepts and binary indicators of students' action to model students' changing knowledge state. 
% However, high dependency on accurate knowledge components labeling is the main drawback of it. 
% However, its prediction accuracy highly depends on the accurate labeling of knowledge components, which may not always be available.
% To reducing the dependency on accurate labeling of knowledge components, DKT~\cite{DKT} and its several variants~\cite{yeung2018addressing, Chen0ZP18} apply RNNs on a sequence of students' problem solving records to extract the hidden knowledge.
However, sequential models cannot be directly applied to student performance prediction in interactive online question pools. They rely on the accurate labeling of the questions' tested knowledge, which is not always available in online question pools. Also, these models do not distinguish different questions of the same knowledge label
% and are not always 
and will predict the same result for different questions with the same knowledge label.
% However, these prior studies cannot be directly applied to student performance prediction in interactive online question pools due to the lack of both accurate knowledge component labelling and fixed sequence of learning materials that are often predefined by the course curriculum.

A recent study~\cite{Wei2020lak} conducted student performance prediction in interactive online question pools by introducing new features based on student mouse movement interactions to delineate the similarity between questions.
However, their approach implicitly requires that the questions must have similar question structure designs and involve drag-and-drop mouse interactions, which may not always hold.
In this paper, we aim to propose a more general approach for student performance prediction in interactive online question pools that can work for question pools with several hundred or thousand questions of different types. 
\section{Background}
\label{sec_context}

% In this study, we work closely with 
% % our industry collaborator 
% Trumptech, a leading educational technology company in Hong Kong, and our experiments are mainly conducted on the dataset collected from their interactive online question pool. 
% This section will briefly introduce the interactive online question pool and the collected data.

% In this part, we will introduce the interactive online question pool which we cooperate with and the collected data.

% \subsection{Interactive online question pool}
% During the study, we have worked closely with a Hong Kong leading educational technology company, Trumptech on their web-based platform\footnote{\href{https://mad.learnlex.com/}{\url{https://mad.learnlex.com/}}} which has provided more than 40000 K-12 students with around 1700 interactive math questions since 2017. 

In this study, our data is collected from \haotian{Learnlex}, an interactive online question pool developed by Trumptech, a leading educational technology company in Hong Kong. This platform contains around 1,700 interactive math questions and has served more than \haotian{100,000} K-12 students \haotian{during the last decade}. 
% These math questions are designed for K-12 students.
% Each question is designed by specialists and assigned three labels, \textit{grade}, \textit{difficulty} and \textit{math dimension}. \textit{Grade} indicates the targeted grade of students and ranges from 0-12. \textit{Difficulty} is an index given by experts in education and question developers for students' reference and it contains five levels (i.e., 1 to 5) representing easy to hard. \textit{Math dimension} is a fuzzy math concept that classifies all questions into 6 classes, numeric, measures, spatial, algebraic, data-handling and geometric. 
Different from questions provided on MOOC platforms, these interactive questions could be freely browsed and answered by students without predefined orders and are merely assigned fuzzy labels, \textit{grade}, \textit{difficulty}, and \textit{math dimension}. \textit{Grade} indicates the targeted grade of students and ranges from 0 to 12. 
\haotian{\textit{Difficulty} is an index of five difficulty levels (i.e., 1 to 5)}. 
% that represent \textit{easy} to \textit{hard} and are assigned by question developers. 
\textit{Math dimension} is a fuzzy math concept indicating the knowledge tested in that question.

\begin{figure}[h!]
\includegraphics[width=\linewidth ]{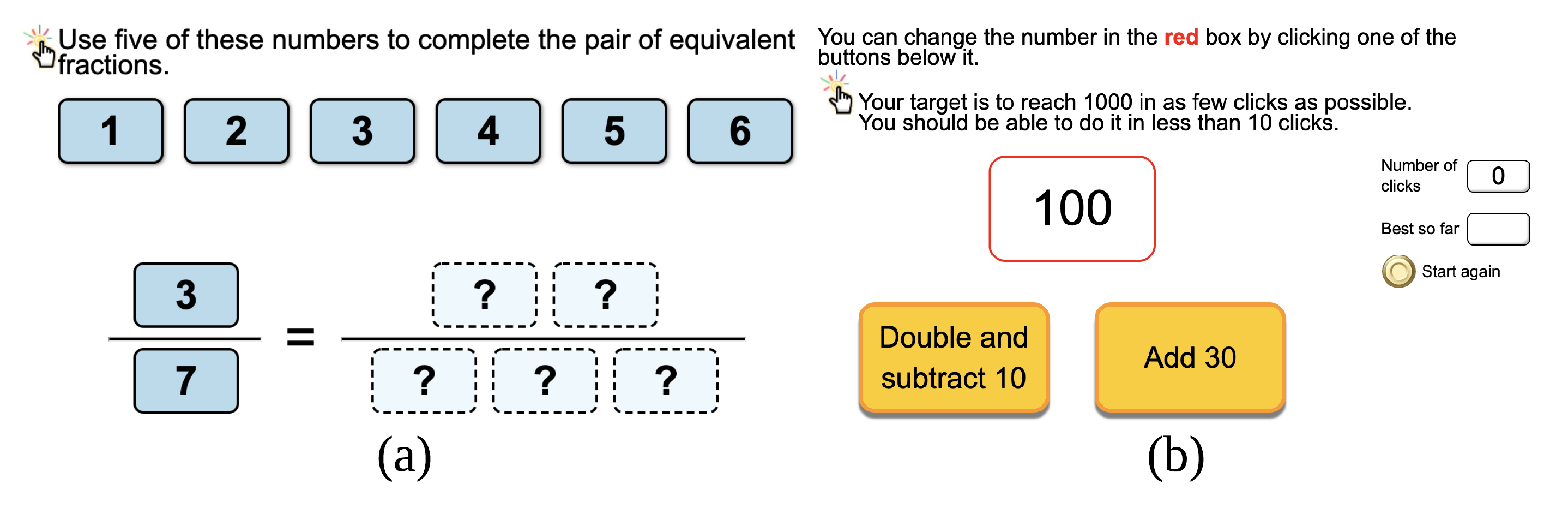}
% \begin{subfigure}{0.48\linewidth}
% \centering
% \includegraphics[width=\linewidth ]{src/figures/Drag-and-drop-example.png} 
% % \vspace{-1em}
% \caption{}
% \label{fig:subim1}
% \end{subfigure}\hfill
% \begin{subfigure}{0.48\linewidth}
% % \centering
% % \vspace{1em}
% \includegraphics[width=\linewidth ]{src/figures/Click-example.png}
% % \hfill
% % \vspace{-1em}
% \caption{}
\vspace{-3em}
% \label{fig:subim2}
% \end{subfigure}
 
\caption{Two examples of interactive questions on \haotian{Learnlex}.}
\vspace{-1em}
\label{example}
\end{figure}

Apart from these labels, the mouse movement interactions of students in their problem-solving process are also collected. 
According to our empirical observation, there are mainly two types of mouse movement interactions during students' problem-solving processes, i.e., \textit{drag-and-drop} and \textit{click}, as shown in Figure~\ref{example}.
% Figure~\ref{example} shows two example questions from this interactive online question pool that require different types of interactions.
Figure~\ref{example}(a) is an example question that needs students to drag blue blocks on the top to appropriate locations to fulfill the requirement (\textit{drag-and-drop}). The question in Figure 1(b) asks students to click the yellow buttons to complete a given task (\textit{click}).

When a student finishes a question, the platform will assign a discrete score between 0 and 100 to the submission. The possible scores of a question are often a fixed number of discrete values depending on what percentages a student can correctly answer the question, and the majority of the questions can have at most four possible score values.
% each question may have different scores,
Therefore, we map the raw scores in historical score records to 4 score levels (0-3) to guarantee a consistent score labeling across questions. Also, only the score of a student's first trial on a question is considered in our experiment.

% After finishing a trial on a question, the system will assign a score ranging from 0-100 and each question can be tried for at most 2 times. Since scores of questions are discrete and each question may have different scores, raw scores in historical score records are mapped to 4 score classes (0-3). Moreover, to reduce noise, we only use the first trials to train and predict the performance. 

% \begin{figure}[h]
%   \begin{minipage}{0.48\linewidth}
%      \centering
%      \includegraphics[width=.9\linewidth]{src/figures/Drag-and-drop-example.png}
%      \caption{Interpolation for Data 1}\label{Fig:Data1}
%   \end{minipage}\hfill
%   \begin{minipage}{0.48\linewidth}
%      \centering
%      \includegraphics[width=.9\linewidth]{src/figures/Click-example.png}
%      \caption{Interpolation for Data 2}\label{Fig:Data2}
%   \end{minipage}
% \end{figure}

% \subsection{Data Collection}
On this platform, we collected 2 parts of data, i.e., the historical score records and the mouse movement records.
There are 973,676 entries from September 13, 2017 to January 7, 2020 in the historical score records, and each entry includes a score value, student ID, question ID, and the timestamp.
% In historical score records, there are 973,676 entries from September 13, 2017 to January 7, 2020, which record the score, student's id, question's id and the timestamp of each submission.
The mouse movement records document the raw types of the mouse events (i.e., \textit{mousemove, mouseup,} and \textit{mousedown}), the corresponding timestamps, and positions of mouse events of all the students working on the interactive online question pool from April 12, 2019 to January 6, 2020. A \textit{mouse trajectory} is a series of raw mouse events that are generated during a student's problem-solving process. In total, we collected 104,113 mouse trajectories made by 4,020 students on 1,617 questions.
\section{The proposed method}

\begin{figure}[h!]
    \includegraphics[width=\linewidth]{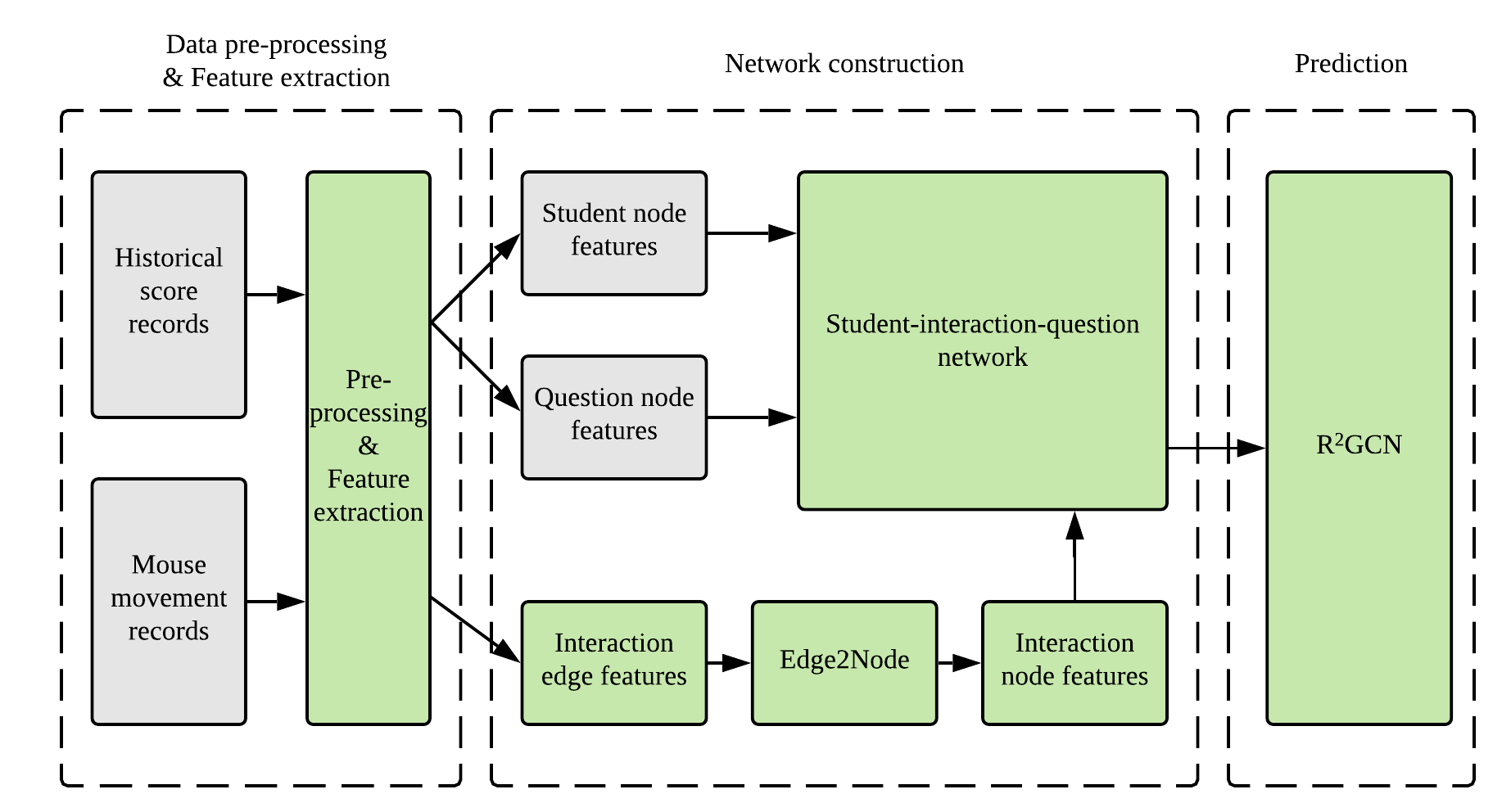}
    \vspace{-2em}
    \caption{The framework of the proposed method. The blocks highlighted in green are our major contributions.}
    % \caption{An Example of interactive question(Area dimension).}
    \label{framework}
    \vspace{-2em}
\end{figure}

We propose a peer-inspired approach for student performance prediction in interactive online question pools. It extensively considers the historical problem-solving records of both a student and his/her peers (i.e., other students working on the question pool) to better model the complex relationship among students, questions, and student performances and further enhance student performance prediction, which is achieved by using a GNN-based model. 
Figure \ref{framework} shows the framework of our approach, which consists of three major modules: data processing $\&$ feature extraction, network construction, and prediction.
% Our framework contains 3 main modules shown in Figure \ref{framework}. We propose this framework to solve the problem of prediction a student's performance on questions requiring either clicks or drag-and-drops to solve in an interactive online question pool. 
The module of \textit{data processing $\&$ feature extraction} is designed to process the related data and extract features of the historical data that will be further used for network construction and student performance prediction.
We considered three types of features:
\textit{statistical features of students} reflecting students' past performance, \textit{statistical features of questions} indicating the question popularity and their average scores, and \textit{mouse movement features} representing the characteristics of students' problem-solving behaviors.
% According to their semantic meaning, in our method, the features extracted are categorized into 3 types, statistical features reflecting students' past performance, statistical features of questions' popularity and average scores as well as the mouse movement features which may reveal students' information during students' problem-solving process.
% The first module is designed to collect, process the related data and extract the feature set. According to their semantic meaning, in our method, the features extracted are categorized into 3 types, statistical features reflecting students' past performance, statistical features of questions' popularity and average scores as well as the mouse movement features which may reveal students' information during students' problem-solving process.
The \textit{network construction} module builds a network consisting of both students and questions, where the interactions between them are also considered and further integrated into the network. This network also incorporates the three types of features to 
extensively model the various performance of different students on different questions. 
% In the second module, the main target is to construct a network of questions with students involved. In our method, we construct a student-interaction-question network with transforming all interaction edges to individual nodes (\textit{Edge2Node}).
Finally, the constructed network, along with the extracted features, is input into the \textit{prediction} module, where we propose R$^2$GCN, a novel Residual Relational Graph Neural Network model that is adapted from R-GCN~\cite{Schlichtkrull2017ModelingRD} by adding residual connections to hidden states, to predict a student's score level on the unattempted questions in interactive online question pools.
% In the third module, a GNN model is applied on the network containing labeled question nodes of a certain student and predicts other unlabeled question nodes. We specifically propose R$^2$GCN model which improves the R-GCN model~\cite{Schlichtkrull2017ModelingRD} with residual connections to hidden states and original features. 
% Finally, we evaluate our proposed method and compare it with several baseline models including traditional machine learning models and GNN models to illustrate the efficiency of our method.

\subsection{Feature Extraction}\label{feature_extraction}

As discussed above, the feature extraction module mainly extracts three types of features: 
% This section will introduce the feature extraction module in detail. Features in our method include 3 types: 
statistical features of students, statistical features of questions, and mouse movement features. 

Statistical features in Tables~\ref{feature_list2} and ~\ref{feature_list3} are extracted from historical score records. Statistical features of students mainly contain students' past performance on various types of questions to reflect the students' ability on a certain type of questions, for example, the average score of first trials on numeric questions of grade 8 and difficulty 3. Statistical features of questions are extracted to show the popularity and real difficulty level of them, for example, the proportion of trials getting score level 2 on the question. 
% Detailed statistical features of students and questions are listed in Appendix~\ref{appendix:student} and ~\ref{appendix:question}.
\begin{table}[h!]
\caption{Statistical features of students.}
\centering
\footnotesize
\setlength{\aboverulesep}{0.5pt}
\setlength{\belowrulesep}{0.5pt}
\vspace{-1em}
\begin{tabular}{p{1.6cm}p{3.05cm}p{2.8cm}}
% \begin{tabular}{@{}lll@{}}

\toprule
Feature Name & Explanation & Example \\
% \hline\hline
\midrule\midrule
\# Total trials &  \begin{tabular}[c]{@{}l@{}}Number of a student's total trials.\end{tabular}& - \\
% \hline
\midrule
\# 2nd trials & \begin{tabular}[c]{@{}l@{}}Number of a student's 2nd trials.\end{tabular}& - \\
% \hline
\midrule
\begin{tabular}[c]{@{}l@{}}\% Trials in \\{[}math dimension$\times$ \\grade$\times$difficulty{]} \end{tabular}& \begin{tabular}[c]{@{}l@{}}Percentage of trials on questions\\ of certain math dimension, \\grade, and difficulty. \end{tabular}& \begin{tabular}[c]{@{}l@{}}\% Trials on spatial questions \\of grade 5 and difficulty 5.\end{tabular}\\
% \hline
\midrule
\begin{tabular}[c]{@{}l@{}}Mean1stScore in \\{[}math dimension$\times$ \\grade$\times$difficulty{]} \end{tabular}& \begin{tabular}[c]{@{}l@{}}Mean score of 1st trials on\\ questions of certain math \\ dimension, grade, and difficulty. \end{tabular}& \begin{tabular}[c]{@{}l@{}}Mean1stScore on numeric\\ questions of grade 8 and\\ difficulty 3.\end{tabular}\\
\bottomrule
\end{tabular}
\label{feature_list2}
\vspace{-1em}
\end{table}

\begin{table}[h!]
%\caption{Statistical features of questions. Features with * are categorical features encoded by one-hot encoding.}
\caption{Statistical features of questions. The star sign * indicates categorical features encoded by one-hot encoding.}
\centering
\setlength{\aboverulesep}{0.5pt}
\setlength{\belowrulesep}{0.5pt}
\footnotesize
\vspace{-1em}
% \begin{tabular}{p{1.8cm}p{3.2cm}p{2.2cm}}
\begin{tabular}{@{}lll@{}}
\toprule
Feature Name & Explanation & Example \\
% \hline\hline
\midrule\midrule
\begin{tabular}[c]{@{}l@{}}Math dimension* \end{tabular} & \begin{tabular}[c]{@{}l@{}} Question's related topic. \end{tabular} & Numeric \\
% \hline
\midrule

Grade* & Grade of target students. & 12 \\
% \hline
\midrule
Difficulty* & \begin{tabular}[c]{@{}l@{}}Question's difficulty level.\end{tabular}& 4 \\
% \hline
\midrule
\# Total trials &  \begin{tabular}[c]{@{}l@{}} Number of trials on a question.\end{tabular}& - \\
% \hline
\midrule
\# 2nd trials & \begin{tabular}[c]{@{}l@{}}Number of 2nd trials on a question.\end{tabular}& - \\
% \hline
\midrule
\begin{tabular}[c]{@{}l@{}}\% Trials in {[}score level{]} \end{tabular}& \begin{tabular}[c]{@{}l@{}}Percentage of trials in each score level. \end{tabular}& \% Trials in 2.\\
\bottomrule
\end{tabular}
\vspace{-1em}
\label{feature_list3}
\end{table}

\begin{figure}[h]
    \centering
    \includegraphics[width=0.8\linewidth]{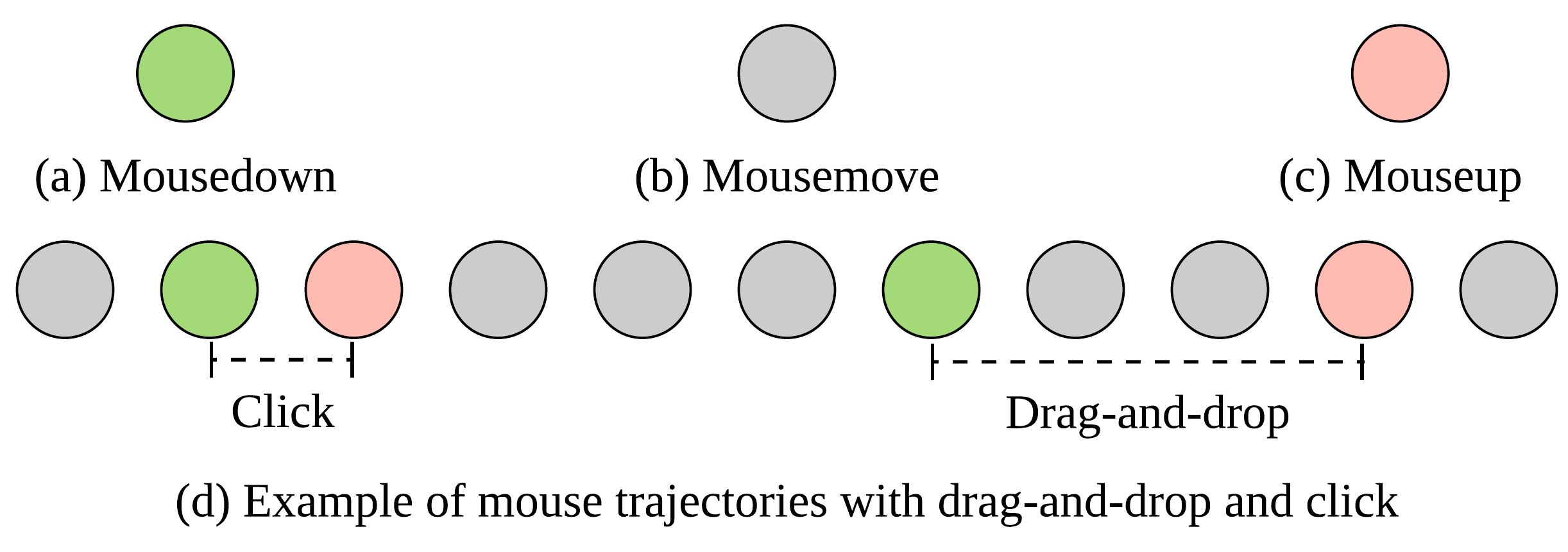}
    \vspace{-1em}
    \caption{An illustration of the mouse interactions \textit{click} and \textit{drag-and-drop}. (a)-(c) indicate raw mouse events.}
    % \caption{An Example of interactive question(Area dimension).}
    \label{clickdrag}
    \vspace{-2em}
\end{figure}

For mouse movement features, we mainly consider two types of basic mouse movement interactions in interactive online question pools: \textit{click} and \textit{drag-and-drop}. However,
despite their differences, both of them start with mouseup and end with mousemove, as shown in Figure~\ref{clickdrag}(d). Thus, they are considered as \textit{generalized clicks (GCs)} in this paper.
% consist of the sequence of the basic mouse events like mousedown, mouseup and mousemove, as shown in Figure~\ref{clickdrag}(d). Thus, they are called \textit{generalized clicks (GCs)} in this paper.
We analyze the GCs in students' mouse trajectories in their problem-solving process and further propose a set of new mouse movement features, as shown in Table~\ref{clickdrag}. These features are mainly designed to reflect the first GC made by students when they try to answer the question. First GCs can reveal the information of questions, for example, the required type of mouse movement interaction. They also reflect the problem-solving behaviors of students, for example, read the description first or try to play with the question first. 
Apart from these new mouse movement features, 
some representative features
% related to think time 
regarding \textit{think time}
introduced in prior studies~\cite{Wei2020lak} are also extracted in this paper to comprehensively delineate students' learning behaviors, for example, the length of thinking time before answering the question.

% we also extract some mouse movement features proposed in the work of Wei~\emph{et al.}~\cite{Wei2020lak} and statistical features of students and questions, for example, historical average scores of students and totally finished times of questions. Detailed features can be found in Appendix~\ref{appendix:feature}.

% Mouse movement interactions used in interactive online question pools mainly contain two types, click and drag-and-drop. In Figure~\ref{clickdrag}, a sample mouse trajectory is displayed and we could notice that drag-and-drops could be viewed as clicks with mouse moves inside. Thus, both drag-and-drops and clicks could be considered as \textit{generalized clicks (GCs)}. According to this characteristic, we manage to propose a feature set of GCs which are extracted from students' mouse trajectories. Also, we extract a feature from mouse movement interaction timestamps. These new features are listed in Table~\ref{clickdrag}. 

\vspace{-1em}

\begin{table}[]
    \caption{New mouse movement features based on GCs and mouse movement timestamps. The star sign * indicates categorical features encoded by one-hot encoding.}
    \vspace{-1em}
\footnotesize
\setlength{\aboverulesep}{0.5pt}
\setlength{\belowrulesep}{0.5pt}
\begin{tabular}{@{}lll@{}}
    \toprule
        Feature Name & Explanation \\
    \midrule\midrule
        1stGCTimeLength & \begin{tabular}[c]{@{}l@{}}Time length between entering the question and the 1st GC. \end{tabular} \\ 
    \midrule
        1stGCTimePercent & \begin{tabular}[c]{@{}l@{}}Percentage of the duration of the 1st GC.\end{tabular} \\ 
        \midrule
        1stGCEventStartIdx & \begin{tabular}[c]{@{}l@{}}Number of mouse events before the 1st GC. \end{tabular} \\ 
        \midrule
        1stGCEventPercent & \begin{tabular}[c]{@{}l@{}}Percentage of mouse events before the 1st GC among all. \end{tabular} \\ 
        \midrule
        1stGCEventEndIdx & \begin{tabular}[c]{@{}l@{}}Number of mouse events when the 1st GC ends. \end{tabular} \\ 
    \midrule
        GCCount & \begin{tabular}[c]{@{}l@{}}Total number of GCs. \end{tabular} \\ 
         \midrule
        GCPerSecond & \begin{tabular}[c]{@{}l@{}} Average number of GCs per second.\end{tabular} \\
         \midrule
        AvgTimeBtwGC & \begin{tabular}[c]{@{}l@{}}Average time between GC.\end{tabular} \\
\midrule     
        MedTimeBtwGC & \begin{tabular}[c]{@{}l@{}}Median value of time between GCs.\end{tabular} \\
        \midrule
        StdTimeBtwGC & \begin{tabular}[c]{@{}l@{}}Standard deviation value of time between GCs. \end{tabular}\\
         \midrule
        OverallDistance & \begin{tabular}[c]{@{}l@{}}Total mouse trajectory length.\end{tabular} \\
        \midrule
        InteractionHour* & \begin{tabular}[c]{@{}l@{}}Time point when students solves the problem, e.g., 13:00.\end{tabular} \\
        \bottomrule
    \end{tabular}
    \label{feature}
    \vspace{-2em}
\end{table}

% Besides new mouse movement features, we also extract some mouse movement features proposed in the work of Wei~\emph{et al.}~\cite{Wei2020lak} and statistical features of students and questions, for example, historical average scores of students and totally finished times of questions. Detailed features can be found in Appendix~\ref{appendix:feature}.
\subsection{Network Construction} \label{network}
\begin{figure}[h!]
    \centering
    \includegraphics[width=0.85\linewidth]{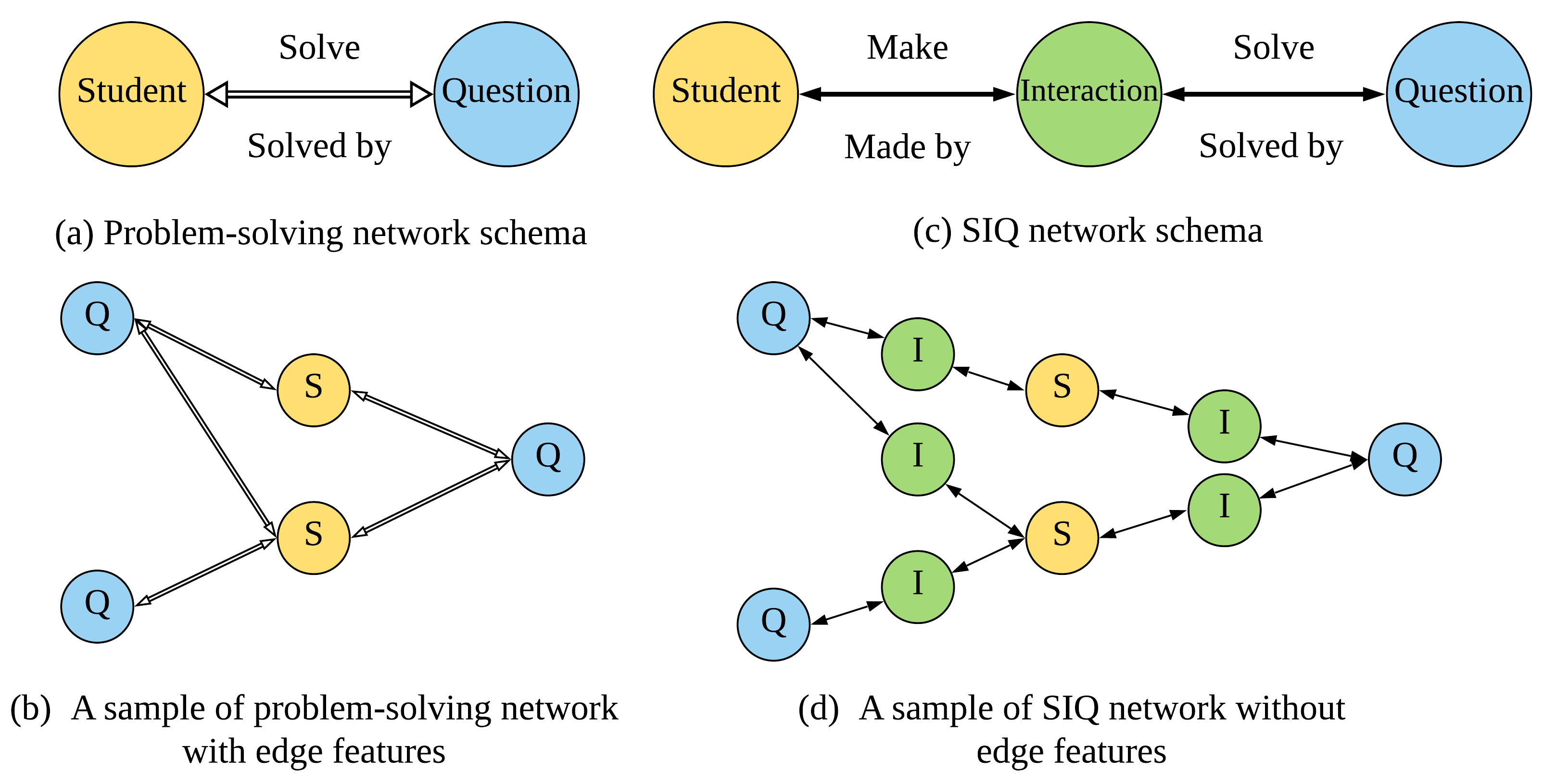}
    \vspace{-1em}
    \caption{\haotian{The problem-solving network and SIQ network.} 
    % The edges in original problem-solving network represent interaction edges with multi-dimensional mouse movement features while the edges in the SIQ network only represent the type of relationship.
    }
    % \caption{An Example of interactive question(Area dimension).}
    \label{fig:network}
     \vspace{-1em}
\end{figure}

It is challenging to model the relationship between questions in an interactive online question pool, since there is no curriculum or predefined question order that every student need to follow and can help model the relationship among questions.
We propose using students' mouse movement interactions with the attempted questions as a bridge to construct the dependency relationship between different questions and build a problem-solving network. When conducting performance prediction for a student, the mouse movement records of his/her peers (i.e., other students) are all considered.
Figure~\ref{fig:network}(a)-(b) illustrates the problem-solving network is a heterogeneous network composed of students nodes (S), questions nodes (Q), and interaction edges with multi-dimensional mouse movement features mentioned above.

% xxxx~\yong{pls briefly describe the figure}. 

% The relationship between, as  questions in an interactive online question pool is not evident since there is no well-designed curriculum or pre-defined question order which could reveal the relationship among questions. Thus, a possible method to solve this problem is to involve peers and their interactions as bridges to build a problem-solving network~\cite{Wei2020lak} in Graph \ref{fig:network}(b). In the network, S and Q denoting Student and Question respectively and one kind of bi-directional edges attached with mouse movement features. 

Inspired by the recent progress of GNNs~\cite{Zhou2018GraphNN}, we propose using GNNs to model the relationship among questions, students, and the interactions between them, which forms intrinsically a heterogeneous network with multi-dimensional edge features. However, there are no GNN models designed for such kind of heterogeneous networks.
Inspired by the method of breaking up edges and transform a homogeneous network to a bipartite network~\cite{Zhou2018GraphNN}, we conduct a transformation named \textit{Edge2Node} to transform the mouse movement interaction (i.e., edges) between students and questions into ``fake nodes'', and further build a Student-Interaction-Question ($SIQ$) Network to model the complex relationships among different questions and students, as shown in Figure~\ref{fig:network}(d).
$SIQ$ network is the basis of applying our GNN-based approach to student performance prediction in interactive online question pools.
% \yong{haotian, pls check my above writing and see if some important ideas are missing. Also, pls check my comments on your original writing.}

% To utilize the graph structures, GNN could be a method to directly perform message passing between nodes along edges. Due to lack of GNN models for the problem-solving network, a heterogeneous network with multi-dimensional edge features, we propose a method to simplify it with Edge2Node transformation inspired by the method of transforming a homogeneous graph with edge information to a bipartite network~\cite{Zhou2018GraphNN} and build our Student-Interaction-Question ($SIQ$) Network to serve as an example. In Edge2Node, we first break up the original solving/solved by edge into two bi-directional edges representing the relationship of make/made by and to solve/solved by as Figure~\ref{fig:network}(d) shows.
% Then we add an interaction node and assign the original mouse movement features attached to interaction edges to it. Other students' and questions' statistical features are still assigned to corresponding nodes.
\subsection{Residual Relational Graph Neural Network} \label{section:gnn}
To model the relationship of questions, students, and interactions, we construct a heterogeneous $SIQ$ network to feed into GNN models. R-GCN~\cite{Schlichtkrull2017ModelingRD} is one of the most widely used models to perform message passing on heterogeneous networks due to its good scalability and excellent performance. 
% However, in original R-GCN, information in hidden states is not fully utilized. 
However, it does not fully make use of the hidden states. In GNNs, hidden states could be considered as the message aggregation results of near neighbors and R-GCN directly transforms them to next hidden states, which leads to a possible information loss.

% \yong{pls insert one short sentence to explain why "not fully utilized"}
% Different level of hidden states are often captured by residual connections in CNNs and the success of introducing residual connections to GCN ~\cite{residual} proves that it also augment the performance of GNNs. 
One of the popular methods to handle this issue is to add residual connections between different level of hidden states and such an approach has been successfully applied to enhancing the performance of GCN~\cite{residual}.
Also, prior research~\cite{wide&deep} has also shown that integrating linear transformation of original simple features to the output layer can help improve the performance of deep models. 
% Further, Wide \& Deep learning~\cite{wide&deep} shows that wide linear models on original simple features helps with deep models. 
Therefore, we propose a new model structure, Residual Relational Graph Convolutional Network (R$^2$GCN) to enhance traditional R-GCN~\cite{Schlichtkrull2017ModelingRD} structure by adding residual connections between different hidden layers and also integrating the original statistical features of questions into the model.

% Wide \& Deep learning~\cite{wide&deep} proposed by Cheng \emph{et al.} introduces a model structure to joint train a wide linear model and a deep neural network to better learn from original features. Dehmamy \emph{et al.} 
% Inspired by Dehmamy's work~\cite{residual}, we propose a new model structure, Residual Relational Graph Convolutional Network (R$^2$GCN), to enhance traditional R-GCN~\cite{Schlichtkrull2017ModelingRD} structure by adding residual connection to hidden states and original statistical features of questions. 

\begin{figure}[h]
    \centering
    \includegraphics[width=0.99\linewidth]{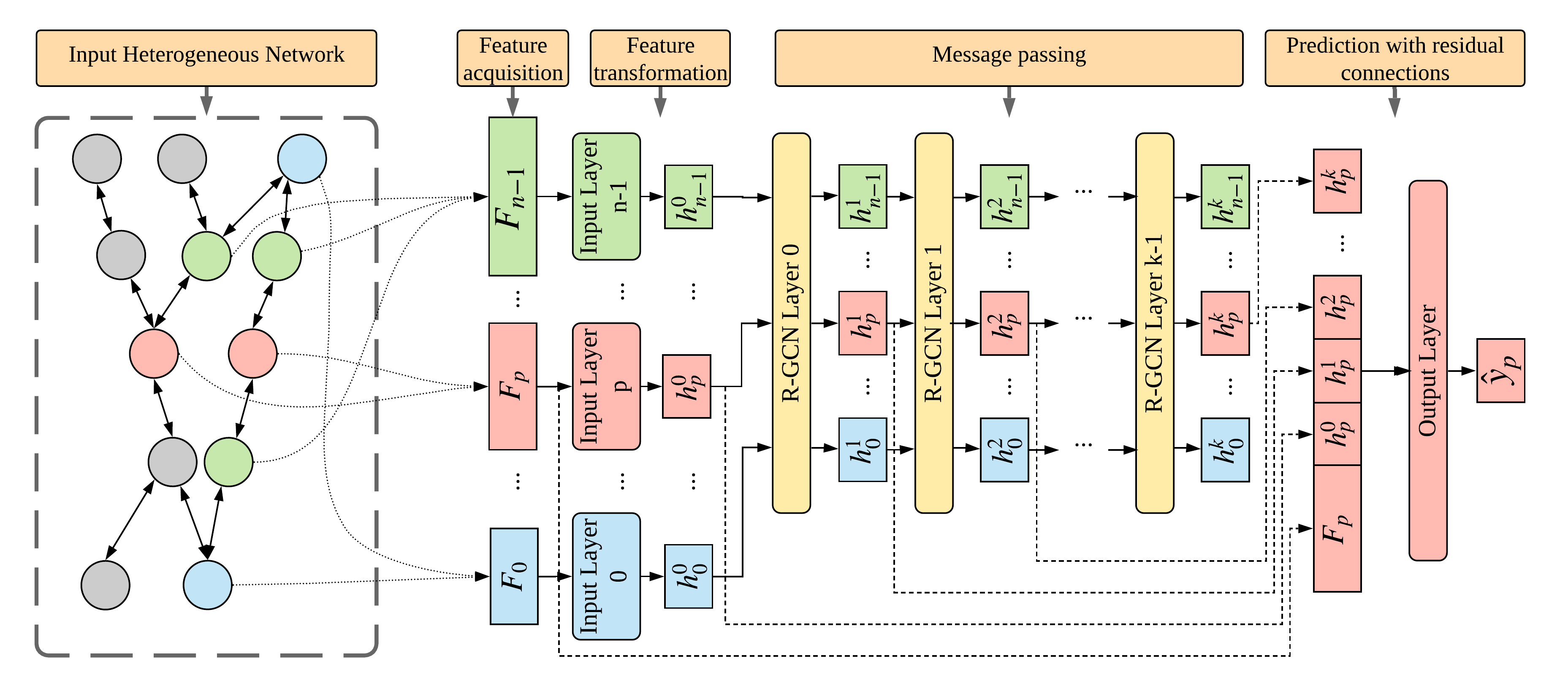}
    \vspace{-1em}
    \caption{The framework of our proposed R$^2$GCN model. The input heterogeneous network contains $n$ types of nodes and nodes of type $p$ is our target for prediction. \haotian{The model has $k$ R-GCN layers. All rounded rectangles represent various layers and other rectangles represent 2D tensors. Dash lines represent residual connections. Each color represents a type of nodes and grey denotes other types of nodes.} $F$ represents features of nodes. $h$ represents hidden states. }
    % \caption{An Example of interactive question(Area dimension).}
    \label{GNN}
    \vspace{-2em}
\end{figure}

Figure~\ref{GNN} shows the framework of the proposed model, R$^2$GCN. 
% The overview of R$^2$GCN is shown in Figure~\ref{GNN} and 
It consists of parallel input layers for feature transformation of different types of nodes to the same shape, consequential R-GCN layers for message passing, residual connections to hidden states and original features for capturing different levels of information and the output layer for final prediction. 
% Following the framework of GNN introduced by Gilmer \emph{et al.}~\cite{Gilmer2017NeuralMP}, 
We will introduce the structure of our model from the perspectives of several key functions, message and update function in the message passing phase and readout function in the prediction phase.
In the input network, $n$, $p$, and $R$ represents the number of node types, the target type of nodes, and all relation types respectively. Since the $SIQ$ network contains three types of nodes, $n$ is 3 in our experiments.

% \subsubsection{Message function} 
% The function transmiting messages from neighbours.  
% \begin{equation} \label{message_func}
% m_{i, j}^{(l+1)}=W_{r}^{(l)} h_{j}^{(l)}
% \end{equation}
% where $W_{r}$ is the weight matrix of relation $r$ and $h_{j}^{(l)}$ is the hidden state of node $j$ after layer $l$. This could also be viewed as a linear transformation of the hidden state of node $j$ and use its result as the message $m_{i, j}^{(l+1)}$ to node $i$.
\vspace{-5pt}
\paragraph{Message function}
% The function transmitting and aggregating messages from all neighbor nodes $\mathcal{N}_{i}$ to center node $i$. 
The message function transmits and aggregates messages from all neighbor nodes $\mathcal{N}_{i}$ to center node $i$ in the message passing phase.
% Since our model distinguishes different kinds of edges, it aggregates neighbor nodes connected by same type of edges first and then aggregates messages from different types of edges. 
In each R-GCN layer, the received message $M_{i}^{(l+1)}$ of node $i$ in layer $l+1$ is defined as 
\vspace{-0.5em}
\begin{equation} \label{reduce_func}
% \vspace{-0.5em}
\textstyle 
M^{(l+1)}_{i} = \sum_{r \in R} \sum_{j \in \mathcal{N}_{i}^{r}}w_{i,j}W_{r}^{(l)} h_{j}^{(l)},
\end{equation}
where $W_{r}$ is the weight matrix of relation $r$ belonging to all relations $R$ and $h_{j}^{(l)}$ is the hidden state of node $j$ after layer $l$, and $w_{i,j}$ indicates the weight of the message from node $j$. Here we use average function to reduce the messages transmitting on the same type of edges and sum function to reduce messages of different types of edges. $w_{i,j}$ is set as the multiplicative inverse of the number of nodes in $\mathcal{N}_{i}^{r}$ in this paper.
\vspace{-5pt}
\paragraph{Update function}
The update function updates the center node $i$'s hidden state $h_{i}^{(l)}$ after layer $l$ with the message $M^{(l+1)}_{i}$ generated by Equation (\ref{reduce_func}) in the message passing phase. 
In our model, to preserve the original hidden state of the center node $i$, the update function is defined as 
\vspace{-0.3em}
\begin{equation} 
\textstyle 
\label{update_func}
h_{i}^{(l+1)} = \sigma(M^{(l+1)}_{i} + W_{0}^{(l)}h_{i}^{(l)} + b),
\end{equation}
where $W_{0}$ denotes the weight matrix of center node $i$'s hidden state, $b$ denotes the bias and $\sigma$ is the activation function.
\vspace{-5pt}
\paragraph{Readout function}
The readout function transforms the final hidden state to the prediction result. 
% Our readout function distinguishes R$^2$GCN from the original R-GCN model by introducing residual connections to both hidden states and original features. 
Different from the original R-GCN model, the readout function of R$^2$GCN adds residual connections to both hidden states and original features.
The readout function for node $p_{i}$ of type $p$ is defined as 
\vspace{-0.5em}
\begin{equation} 
\textstyle 
\label{readout_func}
\hat{y_{p_{i}}} = f(Concat(F_{p_{i}}, h_{p_{i}}^{0}, h_{p_{i}}^{1}, ..., h_{p_{i}}^{k})),   
\end{equation}
where $\hat{y_{p_{i}}}$ is the predicted result, $f$ is the function to transform the concatenated hidden states to final result, $F_{p_{i}}$ denotes the original input features of the corresponding question node and $k$ represents the number of consequential R-GCN layers.
% \yong{pls check the above comments.}
\section{Experiment} 
\label{experiment}
We conducted experiments to evaluate the prediction accuracy and weighted F1 score of students' performance on each question and there are 4 question score levels. 
We compared the proposed approach with 5 baseline approaches including 3 classical machine learning models without peers' mouse movement features. 
% \haotian{In our experiment, we followed the leader board comparison recently raised in machine learning \cite{kipf2017semi, Schlichtkrull2017ModelingRD, Gilmer2017NeuralMP} to test the validation and test data for the comparison of performance.}

% on our dataset collected from a K-12 mathematics online question pool. Specifically, we evaluated the prediction accuracy of students' score on each question and compared our approach with 6 baseline approaches.    

% 4-class semi-supervised classification tasks on our datasets and compare the performance of our proposed method with other baseline methods.

\subsection{Data Processing}
%As introduced in Section~\ref{sec_context}, our data is collected from a K-12 mathematics online question pool and it contains mouse trajectories of 4,020 students on 1,617 questions. 
% In our experiment, we extract two datasets, \textit{short-term dataset} using records from April 12, 2019 to June 27, 2019 and a \textit{long-term dataset} using all records. 
In our experiment, we first extract a portion of the original dataset with records from April 12, 2019 to June 27, 2019 (denoted as \textbf{\textit{short-term dataset}}) for extensively evaluating the proposed approach in terms of prediction accuracy, the influence of labeled dataset size and the influence of the topological distance between questions in training, validation, and test set. 
In short-term dataset, there are 43,274 mouse trajectories made by 3,008 students on 1,611 questions. Taking into account that too few labeled data will make it difficult to train the GNN models, we only conducted experiments for the students who have finished at least 70 questions. Therefore, we gained 47 students in total, which are tested in our experiments for the short-term dataset. 
Also, we further use all the records from April 12, 2019 to January 6, 2020 (denoted as \textbf{\textit{long-term dataset}}) to further evaluate the performance of our approach. We extend the range of filtered students to those who have finished at least 20 questions. Thus, there are in total 1,235 students in this dataset.

For each student $s$, we use 70\% of his/her problem-solving records in the early time period as the training set, the next 15\% records as the validation set, and the last 15\% as the test set. 
% These scores are labeled on corresponding question nodes.
\haotian{
% When processing his/her dataset, we need to split 
% the whole dataset into three sets (i.e., training, validation and test sets) with two timestamps,. The timestamp of spliting training and validation sets, $t^s$, is different for different students.
% , which is denoted as $t^s$.
When processing his/her dataset, the timestamp of the split timestamp between training and validation set is recorded as $t_1^s$
and the split timestamp between validation and test set is recorded as $t_2^s$.
% With \haotian{$t_1^s$ and $t_2^s$}, 
Thus, the $SIQ$ network for student $s$ is built with all students' problem-solving records 
% $T_{t_{start},t_1^s}$ 
between April 12, 2019 and $t_1^s$. Each student has a personalized network of different sizes, which helps provide better performance prediction for different students.
% When constructing the $SIQ$ network, we only consider the questions that have been answered once during the period between April 12, 2019 and $t_1^s$.
% The basic statistics of all the $SIQ$ networks of filtered students in short-term dataset and long-term dataset can be found in Table \ref{network_stat}. 
% Their $SIQ$ networks are built with all available students' records between April 12, 2019 and $t_1^s$.
% and could be viewed as an independent dataset. 
% Furthermore,
All the statistical features assigned to student and question nodes in $SIQ$ networks are extracted from records before April 12, 2019. 
Since $t_1^s$ is always later than that date, the leakage of validation and test data in the training process can be avoided}.
% During the construction of networks, we remove questions that have never appeared in the $T_{t_{start}-t_1^s}$ since they are not connected in the $SIQ$ network.

% We also remove students who did not appear before April 12, 2019, since they have no valid statistical features.

% \yong{pls carefully link the above sentences and make the logic smooth.}

% \begin{table}[]
% \centering
% \begin{tabular}{lllllll}
% \hline
%  & Type & Mean & Median & Max & Min& Labeled Mean \\ \hline
%  & Student & 2338 & 2529 & 2873 & 782 & 89 \\ \cline{2-7} 
% Node & Interaction & 28962 & 32415 & 42409 & 5608 &- \\ \cline{2-7} 
%  & Question & 1547 & 1593 & 1608 & 1121&-  \\ \hline
%  & Make & 28962 & 32415 & 42409 & 5608&-  \\ \cline{2-7} 
% \multirow{2}{*}{Edge} & Made by & 28962 & 32415 & 42409 & 5608&-  \\ \cline{2-7} 
%  & To solve & 28962 & 32415 & 42409 & 5608&-  \\ \cline{2-7} 
%  & Solved by & 28962 & 32415 & 42409 & 5608&-  \\ \hline
% \end{tabular}
% \caption{This table shows the basic statistics of all 47 students' SIQ networks in our experiment. All kinds of edges and interaction nodes share the same values since they are generated together from the original solving/solved by edges in problem-solving networks.}
% \label{network_stat}
% \end{table}

\vspace*{-5pt}
\subsection{Baselines}\label{Baselines}
We compared our approach with 
both the state-of-the-art GNN models and other traditional machine learning approaches 
% various approaches
for student performance prediction to extensively evaluate the performance of our approach. 
These baselines are as follows:
% Baseline models used to extensively evaluate the performance of our approach are as follows:

\textbf{\textit{R-GCN}}: a classical GNN model proposed for networks with various types of edges. 
% We test R-GCN as well as two deteriorated variants of our model $R^2GCN$. 
We test R-GCN with two variants of our networks.
R-GCN (without E2N) represents the input of R-GCN model is a problem-solving network in Figure \ref{network}(b) without edge features. R-GCN (with E2N) denotes the input is the $SIQ$ network with $Edge2Node$ transformation.
% \paragraph{GCN~\cite{kipf2017semi}}
% A GNN model proposed for graphs with one type of nodes and one type of edges. Since it is not able to handle multiple relations and various node types,we use the similarity calculation method proposed by Wei \emph{et al.}~\cite{Wei2020lak} to eliminate student nodes and solving/solved by edges in problem-solving network and use the similarity scores as edge weights.
% \vspace*{-5pt}

\textit{\textbf{GBDT}}: a tree model utilizing the ensemble of trees.
To verify the effectiveness of integrating peer information into student performance prediction in our approach, we only consider statistical features of students and questions in Tables~\ref{feature_list2} and \ref{feature_list3} in GBDT. 

% To prove the effectiveness of our peer-inspired model, we do not involve peers and their mouse movement features and only feed statistical features of students and questions into GBDT. 

% \vspace*{-5pt}

\textit{\textbf{SVM}}: a model constructing a hyperplane or hyperplanes to distinguish samples. Similar to GBDT, only statistical features of students and questions in Tables~\ref{feature_list2} and \ref{feature_list3} are fed into SVM.
% The same statistical features as GBDT is fed.
% \vspace*{-5pt}

\textit{\textbf{LR}}:
a classical linear model with a logistic function to model dependent variables.
We use features in Tables~\ref{feature_list2} and \ref{feature_list3} for LR.
% The same statistical features as GBDT and SVM is fed.
% \subsection{Model settings, training parameters and running environment}
\subsection{Detailed Implementation}
Our GNN models are mainly implemented using PyTorch
% \footnote{\href{https://pytorch.org/}{https://pytorch.org/}} a
and DGL\footnote{\href{https://www.dgl.ai/}{https://www.dgl.ai/}}, while GBDT, LR, and SVM are implemented with Sci-kit Learn.
% \footnote{\href{https://scikit-learn.org/stable/}{https://scikit-learn.org/stable/}}.
% In the R$^2$GCN model used in our experiment, there are 3 parallel input layers in Figure \ref{GNN} to transform original node features of 3 types of nodes. 
For our model R$^2$GCN, we use three parallel input layers to transform original features of 3 types of nodes, as shown in Figure~\ref{GNN}.
Then we use 3 sequential R-GCN layers with a hidden size of 128.
% with 128 hidden units. 
The final two layers of our model are fully-connected neural networks with a hidden size of 128.
% and ReLU is as the activation function.
% neural network. 
The activation function used in our model is ReLU.
% In the training of all GNN models, all models are optimized with Adam~\cite{adam}.
All the GNN-based models in our experiments use Adam as the optimizer and cross entropy as the loss function.
% Loo
We empirically set the learning rate as $1e\mhyphen4$ and weight decay rate as $1e\mhyphen2$. 
The early stopping mechanism is applied to our GNN models. The maximum number of training epochs is set as $400$ and the early stopping patience is set as $100$ epochs. For GBDT, we set the number of trees as $250$, the max depth as 5, and the learning rate as $1e\mhyphen3$. 
For SVM, we use Radial Basis Function (RBF) kernel and the regularization parameter is set as $1$.
% Due to the randomness of machine learning models, 
To gain reliable results,
we trained and tested every model for 10 times and report the average performance.
\begin{table}[h!]
% \caption{ This table shows the basic statistics (mean, max, min value of node number and mean value of labeled nodes) of all students' $SIQ$ networks in \emph{short-term dataset} and \emph{long-term dataset}. All kinds of edges and interaction nodes share the same values since they are generated together from the original solving/solved by edges in problem-solving networks.}
\caption{The basic statistics of all students' $SIQ$ networks. $Q$, $S$, $I$ denote question nodes, student nodes, and interaction nodes respectively. 
Label shows average number of labeled $Q$ nodes in all students' networks.
% \yong{What does this mean?}
}

\setlength{\aboverulesep}{0.5pt}
\setlength{\belowrulesep}{0.5pt}
\footnotesize
\centering
\vspace{-1em}
\begin{tabular}{p{0.6cm}p{0.5cm}p{1.9cm}p{0.6cm}<{\centering}p{0.6cm}<{\centering}p{0.5cm}<{\centering}p{0.6cm}<{\centering} }
\toprule

& & Type & Mean & Max & Min& Label\\ 
% \hline\hline
\midrule\midrule
\multirow{4}{*}{\makecell{Short- \\ term}} & & S & 2,338  & 2,873 & 782 & - \\ 
% \cline{3-7}
\cmidrule{3-7}
& Node & I & 28,962  & 42,409 & 5,608 &- \\ 
% \cline{3-7} 
\cmidrule{3-7}
& & Q & 1,547  & 1,608 & 1,121&89  \\ 
% \cline{2-7} 
\cmidrule{2-7}
&Edge
& $I$-$Q$, $Q$-$I$, $I$-$S$ \& $S$-$I$ & 28,962  & 42,409 & 5,608&-  \\ 
% \cline{3-7}
% \cmidrule{3-7}

% \hline
\midrule
\multirow{4}{*}{\makecell{Long- \\ term}} & & S & 2,749  & 3,556 & 89 & - \\ 
% \cline{3-7}
\cmidrule{3-7}
& Node & I & 52,485  & 92,573 & 450 &- \\ 
% \cline{3-7}
\cmidrule{3-7}
& & Q & 1,540  & 1,597 & 419&58  \\ 
% \cline{2-7}
\cmidrule{2-7}
&Edge
& $I$-$Q$, $Q$-$I$, $I$-$S$ \& $S$-$I$ & 52,485  & 92,573 & 450&-  \\ 
% \cline{3-7} 
% \cmidrule{3-7}
% &\multirow{2}{*}{$QI$-Edge}
%  & $I$-$Q$& 52485  & 92573 & 450&-  \\ \cline{3-7} 
% & & $Q$-$I$& 52485  & 92573 & 450&-  \\ \cline{2-7}
 
% & \multirow{2}{*}{$SI$-Edge} 
%  & $I$-$S$& 52485  & 92573 & 450&-  \\ \cline{3-7} 
% & & $S$-$I$& 52485  & 92573 & 450&-  \\
 \bottomrule
\end{tabular}
\label{network_stat}
\vspace{-2em}
\end{table}

\subsection{Evaluation Metrics}
We use three different metrics to evaluate models comprehensively. Here we use $s$, $n_{c}^s$, $n^s$, $W\mhyphen F1^{s}$ to denote a student, the number of correctly predicted questions for a student, the number of questions in the test set, and the weighted F1 of the prediction results.

\textbf{\textit{Average personal accuracy (AP-Acc)}} evaluates a model's average prediction accuracy on different students:
\begin{equation}
\textstyle 
    AP\mhyphen Acc = \frac{1}{N}\times\sum_{s=1}^{N}\frac{n_{c}^s}{n^s}.
\end{equation}

\textbf{\textit{Overall accuracy (O-Acc)}} evaluates a model's average prediction accuracy on all predicted questions:
\begin{equation}
\textstyle 
    O\mhyphen Acc = \sum_{s=1}^{N}{n_{c}^s}/\sum_{i=1}^{N}{n^s}.
\end{equation}

\textbf{\textit{Average personal weighted F1 (APW-F1)}} evaluates a model's average weighted F1 score on different students:
\begin{equation}
\textstyle 
    APW\mhyphen F1 = \frac{1}{N}\times\sum_{s=1}^{N}W\mhyphen F1^{s}.
\end{equation}

\subsection{Short-term Dataset}

% we filtered students who have finished 70 questions during the period in order to have enough data for the experiment in Section~\ref{size}.
% Total number of filtered students is 47. 

\subsubsection{Prediction accuracy}
% In our experiments, we compare our proposed methods with baselines on accuracy and weighted F1 scores. 
% We first compare our approach with the baseline methods discussed in Section~\ref{Baselines} in terms of the accuracy of student performance prediction.
% Three different metrics are used in the evaluation of accuracy, i.e., average personal accuracy (AP-Acc), overall accuracy (O-Acc) and average personal weighted F1 (APW-F1).
% AP-Acc is the average personal accuracy of all the 47 students using a prediction method and APW-F1 is the average personal weighted F1 scores of of all the 47 students' models. 
% Also, since the number of every student's labeled nodes may not always be the same, 
% we further calculate O-Acc as
% \begin{equation}
%     O\mhyphen Acc = \frac{\sum_{s=1}^{N}{n_{c}^s}}{\sum_{i=1}^{N}{n^s}}
% \end{equation}
% The detailed calculation method of our three metrics are illustrated in Appendix ~\ref{evaluation}.

% the overall accuracy on all test samples regardless of students in short-term dataset. 
% we evaluate our method with O-Acc which calculates the accuracy on labeled data of all students. 
\haotian{In our experiments, we follow the prior research on Graph Neural Networks~\cite{kipf2017semi, Schlichtkrull2017ModelingRD, Gilmer2017NeuralMP} and compare the model prediction accuracy of our approach with that of other baseline approaches on the test data.}
Table~\ref{result_baseline} shows the results of our experiments.
Among all the methods, R$^2$GCN performs best across different metrics, which demonstrates the effectiveness of our proposed model. 
% \haotian{It outperforms our best baseline method, R-GCN (with E2N), by 5.4\% in AP$\mhyphen$Acc, by 5.2\% in O$\mhyphen$Acc and by 7.2\% in APW$\mhyphen$F1.}
Also, the peer-inspired GNN models (i.e., R$^2$GCN, R-GCN) outperform all traditional machine learning models, which confirms the advantages of our peer-inspired student performance prediction for interactive online question pools. Moreover, the comparison of R-GCN (with E2N) and R-GCN (without E2N) could justify that using the same model, our approach of constructing the $SIQ$ network outperforms the original problem-solving network.
% It proves that our method advances the performance prediction in online interactive question pools. 
% Another finding in our results is that our models could achieve very close AP-Acc and O-Acc, which indicates that our method can achieve a more consistent student performance prediction result among all the 47 students who have finished at least 70 questions than other traditional machine learning methods.

% that when number of total questions exceed 70, the performance is consistent among all students regardless of the numbers of solved questions.

\begin{figure}[h!]
    \centering
    \includegraphics[width=0.85\linewidth]{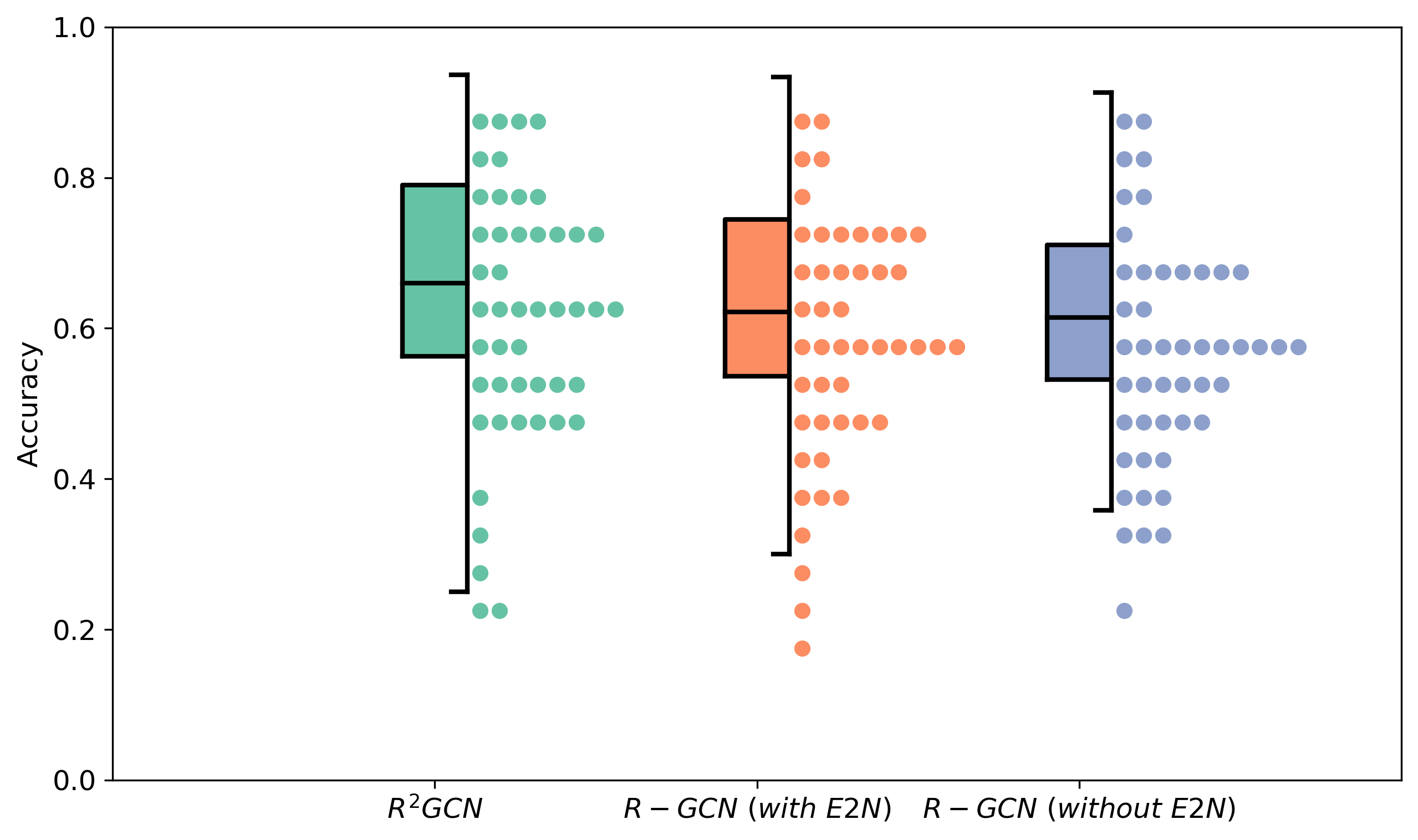}
    \vspace{-1em}
    \caption{The prediction result of each student with three GNN-based methods, R$^2$GCN, R-GCN (with E2N), and R-GCN (without E2N). The dots on the right of the box plots represent students and their vertical positions denote the approximate accuracy. 
    % This figure shows that our proposed method achieve the best performance and have less extremely low cases.
    } 
    % \yong{pls check my comments. You still need to update the figure.}}
    % \caption{An Example of interactive question(Area dimension).}
    \label{box}
    \vspace{-1.5em}
\end{figure}

Figure \ref{box} further shows the box plots with a beam display to show the detailed accuracy distribution of all the 47 students. The box plots indicate that our method can achieve the highest median accuracy. Moreover, from the distribution of personal accuracy on the right of each box, we can learn that our R$^2$GCN has the least number of students whose accuracy is lower than $0.4$, which indicates that extremely lower cases are fewer.

% To further show the accuracy of all students' model, box plots with beam display are drawn to compare results of GNN models in Figure \ref{box}. The plots show that our method can achieve the highest median accuracy and the variation of results is acceptable.

\begin{table}[h!]
%\caption{The result comparison of our proposed R$^2$GCN and baseline models on \emph{short-term dataset}. 
\caption{The comparison between our method R$^2$GCN and the baseline models on \emph{short-term dataset}. 
% E2N represents that interaction features and $Edge2Node$ transformation are used in that model. AP-Acc is the average personal accuracy, O-Acc is the overall accuracy and APW-F1 is the average personal weighted F1 score. 
}
\small
\setlength{\aboverulesep}{0.5pt}
\setlength{\belowrulesep}{0.5pt}
\centering
\vspace{-1em}
\begin{tabular}{p{3.5cm}llllp{2.5cm}}
\toprule
Model & AP-Acc & O-Acc & APW-F1 \\ 
% \hline\hline
\midrule\midrule
R$^2$GCN& \textbf{0.6642} & \textbf{0.6662} & \textbf{0.6148} \\ 
% \hline
\midrule
R-GCN (with E2N) & 0.6302 & 0.6331 & 0.5737 \\
% \hline
\midrule
R-GCN (without E2N) & 0.6151 & 0.6198 & 0.5508 \\ 
% \hline
\midrule
GBDT & 0.5687 & 0.5750 & 0.4398 \\ 
% \hline
\midrule
SVM & 0.5734 & 0.5805 & 0.4470 \\ 
% \hline
\midrule
LR & 0.5928 & 0.5961 & 0.5414 \\ 

\bottomrule
\end{tabular}
\vspace{-1em}
\label{result_baseline}
\end{table}

\subsubsection{Size of labeled data}
\label{size}
\begin{figure}[h!]
    \centering
    \includegraphics[width=0.85\linewidth]{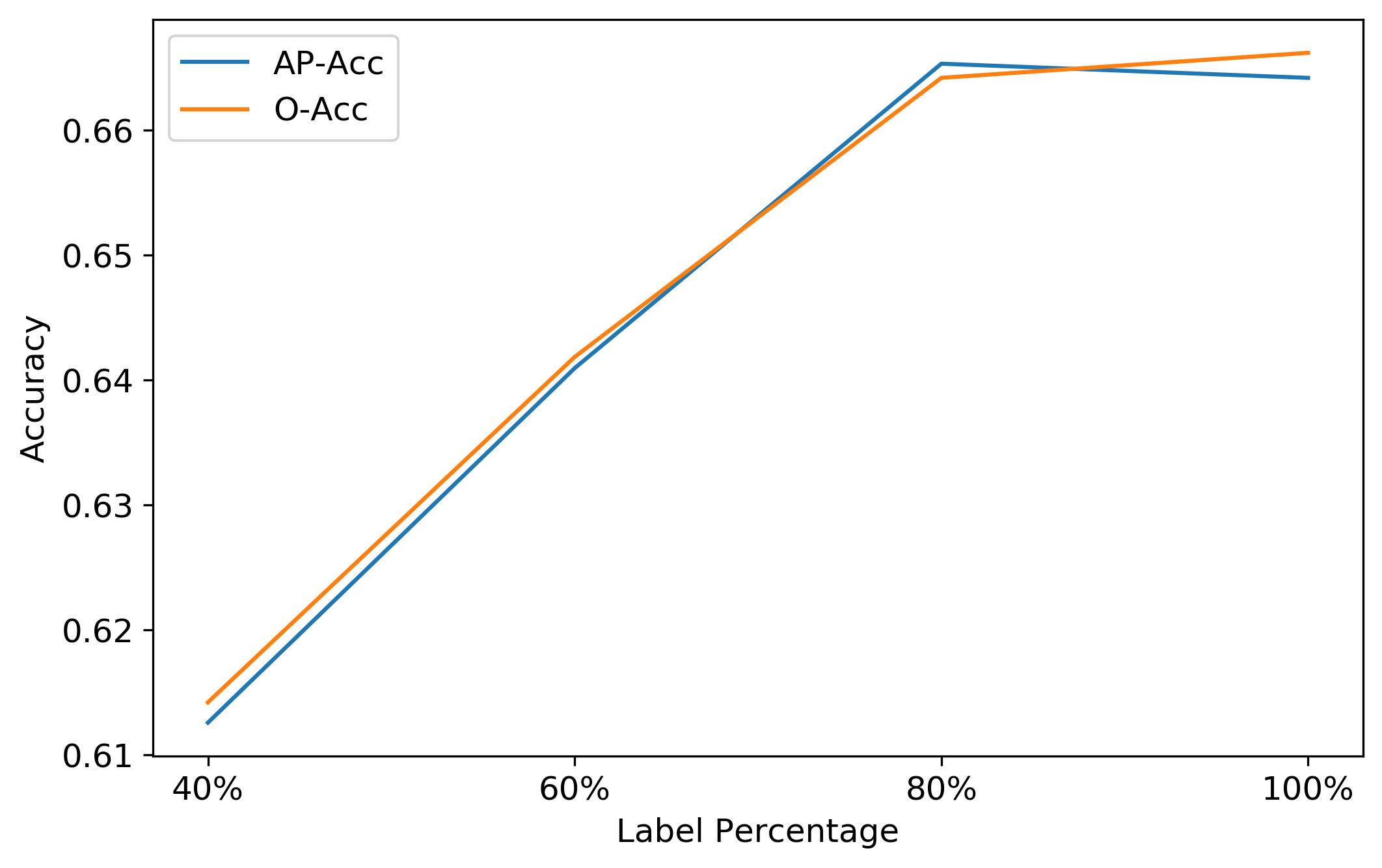}
    \vspace{-1em}
    \caption{The change of AP-Acc and O-Acc along with the change of size of training labels.}
    % \caption{An Example of interactive question(Area dimension).}
    \label{result}
    \vspace{-1.5em}
\end{figure}

% To further evaluate the performance of our method and show the relationship between prediction accuracy and number of training labels, we conducted the experiment on the size of training labels.
We further investigate the influence of the training data size on the final prediction accuracy.
To maintain the consistency of network structure, test set, and validation set, we choose to keep 40\%, 60\%, and 80\% of records in the training set to conduct this experiment. The experiment results of R$^2$GCN is shown in Figure~\ref{result}. It is easy to find that the prediction accuracy increases with the growth in training dataset size and finally reaches a relatively stable prediction accuracy around $0.66$. 
This result is reasonable since there will be more ground-truth labels to guide the training of GNNs when more questions are finished.
% when the size of training labels is growing, the performance is also growing and finally becomes stable. This result indicates that our prediction performance could be better when the student finish more questions. In practice, we may need to use at least around 40 training labels of a student to get high and stable prediction results according to this experiment.

% \vspace{-1em}
% In Figure \ref{result}, we could find that when the size of training labels is growing, the performance is also growing and finally becomes stable. This result indicates that our prediction performance could be better when the student finish more questions. In practice, we may need to use at least around 40 training labels of a student to get high and stable prediction results according to this experiment.

\subsubsection{Topological distance among training, validation, and test set}
\label{experiment:topological}

Apart from the number of training labels, the student performance prediction can also be influenced by the topological distance between the test 
set, and the training or validation set.
% In order to provide some insights in our method, we conduct some visual analysis on the relationship of some topological features and the prediction result. 
Thus, we further calculate the average shortest distance in the $SIQ$ network between questions in training set, test set, and validation set. These average distances are represented by $\overline{d}_{(train, test)}$, $\overline{d}_{(train, val)}$, and $\overline{d}_{(test, val)}$ respectively. Since the interaction nodes are derived from interaction edges, to simplify our analysis, we remove those nodes and use the problem-solving network in Figure~\ref{network} to calculate the shortest path distance with \haotian{NetworkX}.
% \footnote{\href{https://networkx.github.io/documentation/stable/index.html}{https://networkx.github.io/documentation/stable/index.html}}. 
The average shortest distance is calculated as follows:
% \vspace{-0.4em}
\begin{equation}
\textstyle 
    \overline{d}_{(\mathbf{X}, \mathbf{Y})} = \frac{1}{\left \| \mathbf{X} \right \|\left \| \mathbf{Y} \right \|}\sum_{i \in \left \| \mathbf{X} \right \|} \sum_{j \in \left \| \mathbf{Y} \right \|} d_{(x_{i}, y_{j})},
\end{equation}
where $\mathbf{X}$ and $\mathbf{Y}$ denote two sets of questions and $d_{(x_{i}, y_{j})}$ is the shortest path distance between $x_{i}$ and $y_{j}$.

% In Figure \ref{pcp}, we make a parallel coordinates plot\footnote{\href{https://en.wikipedia.org/wiki/Parallel\_coordinates}{https://en.wikipedia.org/wiki/Parallel\_coordinates}} to show the relationship between average distances and personal validation and test accuracy.

\begin{figure}[h]
    \centering
    \includegraphics[width=0.85\linewidth, ]{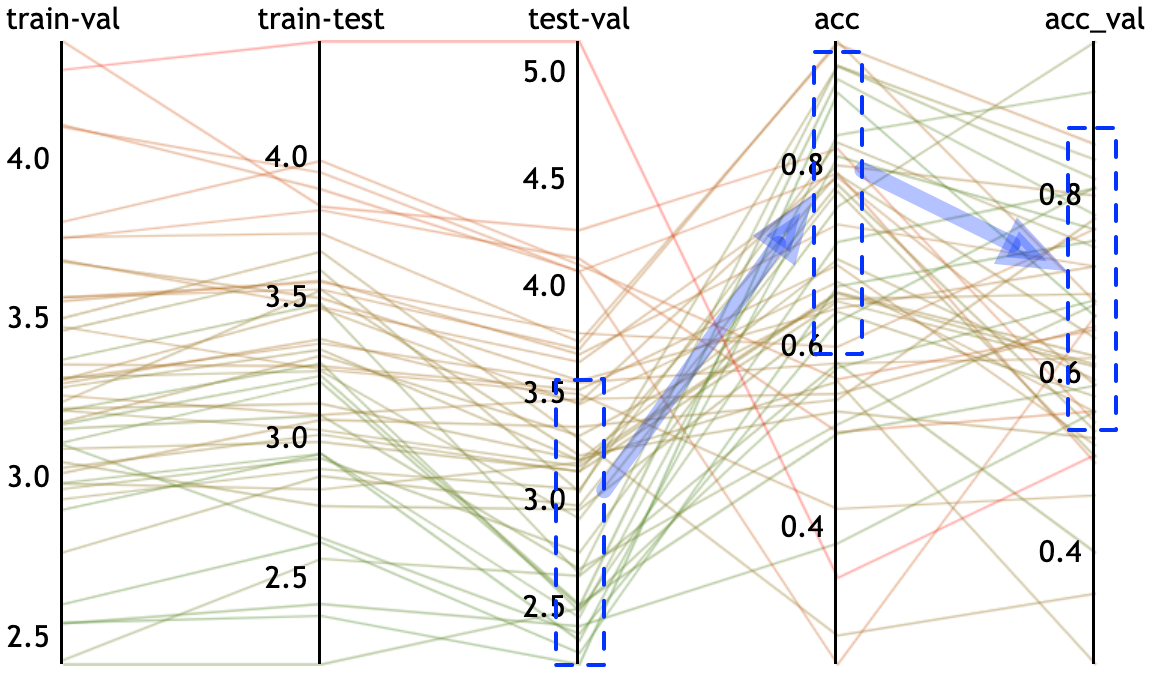}
%    \caption{The relationship between topological distance among training, validation, and test set and AP-Acc. The first 3 parallel y-axes represent $\overline{d}_{(train, val)}$, $\overline{d}_{(train, test)}$, and $\overline{d}_{(test, val)}$ of each students and the last 2 parallel y-axes denote the test AP-Acc (acc) and validation AP-Acc (acc\_val) respectively. Each line represents the data of one student in our dataset and the color of lines maps large $\overline{d}_{(test, val)}$ to red and small $\overline{d}_{(test, val)}$ to green. 
%    % From the plot we could notice the negative correlation of $\overline{d}_{(test, val)}$ and test accuracy and the negative correlation of test and validation accuracy when $\overline{d}_{(test, val)}$ is large.
%    }
    \caption{ The influence of topological distance among training, validation, and test set on the prediction accuracy.
    The first three axes represent $\overline{d}_{(train, val)}$, $\overline{d}_{(train, test)}$, and $\overline{d}_{(test, val)}$ of each student. The last two axes denote the average personal accuracy of each student on their test and validation set, respectively. Each line represents the data of one student and the line color indicates the value of $\overline{d}_{(test, val)}$ (the green to red scheme denotes the small to large values).
%    and the color of lines maps large $\overline{d}_{(test, val)}$ to red and small $\overline{d}_{(test, val)}$ to green. 
	% From the plot we could notice the negative correlation of $\overline{d}_{(test, val)}$ and test accuracy and the negative correlation of test and validation accuracy when $\overline{d}_{(test, val)}$ is large.
}
    \label{pcp}
    \vspace{-1em}
\end{figure}

We use parallel coordinates plot (PCP)
% ~\footnote{\href{https://en.wikipedia.org/wiki/Parallel\_coordinates}{https://en.wikipedia.org/wiki/Parallel\_coordinates}} 
to show the influence of the average distances on the student performance prediction accuracy, as shown in Figure \ref{pcp}. We use 5 parallel y-axes to encode the three average distances and two accuracy scores (i.e., test accuracy and validation accuracy), respectively. 
Each line represents a student in the dataset. 
Here we also use a green-to-red color scheme to encode $\overline{d}_{(test, val)}$ with green indicating lower $\overline{d}_{(test, val)}$ and red indicating high $\overline{d}_{(test, val)}$. 
% \yong{pls check if the above statement is correct.}
It is easy to recognize that there is a negative correlation between the average distance from test to validation set $\overline{d}_{(test, val)}$ and the  accuracy $acc$.
% The negative correlation of $\overline{d}_{test-val}$ and test accuracy is easy to notice. 
Also, we could notice the test accuracy of students with a larger $\overline{d}_{(test, val)}$ is usually lower than their validation accuracy.
% This phenomenon could be explained by that the large distance between questions in the test set and the validation set may lead to the dissimilarity of two sets.
Such an observation is probably because a large average distance between questions in the test set and questions in the validation set indicates the dissimilarity between them, making the early stopping point not always the best point of achieving the best test accuracy. 
% \yong{pls double check my explanation.}
% Thus the early stopping point is not the point of best test accuracy.
% However, most of the students' $\overline{d}_{(test, val)}$ is small, which confirms the usefulness of utilizing the early stopping mechanism to avoid overfitting during the training of the GNN model.
% % so it is necessary to utilize the early stopping mechanism to avoid overfitting the training set. 
% In our experiments, we also found that the overall performance is better when we use an adaptive early stopping mechanism.

\vspace{-0.5em}
\subsection{Long-term Dataset}
To further evaluate the effectiveness and generalizability of our method, we compare the performance with the baseline methods on the long-term dataset that covers the problem-solving records of more students than the short term dataset.
% as shown in Table \ref{result_baseline2}.
The results in Table~\ref{result_baseline2} indicates that when the number of labeled data is limited, our approach can still achieve high accuracy and F1 score.
% \haotian{It is 7.6\% better than R-GCN (without E2N) in AP$\mhyphen$Acc, 6.7\% better than R-GCN (with E2N) in O$\mhyphen$Acc and 9.7\% better than R-GCN (with E2N) in APW$\mhyphen$F1.}
% Also, our approach can achieve a higher O-Acc than AP-Acc on our \emph{long-term dataset}, which implies that the prediction accuracy can be further improved when more questions are finished by more students.
% \yong{pls further check my comments.}

% comparing with AP-Acc, a higher O-Acc on our \emph{long-term dataset} implies that growing number of finished question results in improving accuracy.

\begin{table}[h!]
	\vspace{-1em}
	\caption{The comparison between our method R$^2$GCN and baseline models on \emph{long-term dataset}.}
	\centering
	\small
	\vspace{-1em}
	\setlength{\aboverulesep}{0.5pt}
	\setlength{\belowrulesep}{0.5pt}
	\begin{tabular}{p{3.5cm}llllp{2.5cm}}
		\toprule
		Model & AP-Acc & O-Acc & APW-F1 \\ 
		%  \hline\hline
		\midrule\midrule
		R$^2$GCN& \textbf{0.5507} & \textbf{0.5671} & \textbf{0.5050} \\ 
		% \hline
		\midrule
		R-GCN (with E2N) & 0.5100 & 0.5313 & 0.4605 \\ 
		% \hline
		\midrule
		R-GCN (without E2N) & 0.5119 & 0.5296 & 0.4535 \\ 
		% \hline
		\midrule
		GBDT & 0.4836 & 0.4610 & 0.3686 \\ 
		% \hline
		\midrule
		SVM & 0.4973 & 0.4718 & 0.3801 \\ 
		% \hline
		\midrule
		LR & 0.4881 & 0.4904 & 0.4322 \\ 
		
		\bottomrule
	\end{tabular}
	\vspace{-1.5em}
	\label{result_baseline2}
\end{table}

% \paragraph{Student 2}
% Training accuracy changes of student 2's model is shown in Figure \ref{}. Noticing that the trend of training set and that of test and validation set is not the same - while training accuracy increase, both test accuracy and validation accuracy remain low. This phenomenon reflects that model training on the training set could not fit test and validation set. After checking the average distances, we could notice that $d_{train-val}$ is X.x times above the average $d_{train-val}$ and $d_{train-test}$ is X.x times above the average $d_{train-test}$.

\vspace{-0.5em}
\section{Deployment}
\label{deployment}
% We have been working closely with our industry collaborator Trump-tech for about one year on this research project. 
% When finishing the proposed peer-inspired student performance prediction approach, we interviewed 5 online question designers and course instructors from Trumptech and they showed great interest in our approach and one senior online question designer commented \emph{``the prediction accuracy is impressive compared with other state-of-the-art approaches.
% , especially considering the task is to make a 4-class prediction. 
% We'd like to deploy it on our platforms to facilitate other tasks, e.g., an adaptive question recommendation system''}.
% The proposed peer-inspired student performance prediction approach is going to be deployed on their interactive online K-12 math question pool \emph{LearnLex} soon.
% Currently, we are 
We have been
working with engineers from Trumptech to deploy and further evaluate our proposed peer-inspired student performance prediction approach on their interactive online math question pool\haotian{, \textit{Learnlex}, since June 2020}.
\haotian{By now, our approach has already been integrated into it and is still under active testing, which is expected to be finished in September 2020.}
% After we finish the initial integration of our approach into Learnlex, one senior engineer commented that \emph{``the prediction accuracy is impressive compared with other state-of-the-art approaches. It provides student performance prediction on each question, which can hardly be finished by prior knowledge tracing-based algorithms.''}}
\haotian{
% During the deployment process, one key issue we faced is 
One key issue we have faced by now is
how to handle the \textit{cold start problem}, as our R$^2$GCN model intrinsically requires that there should be some existing training data for newly-registered students on Learnlex. 
% The performance prediction accuracy will be low, when few questions have been solved by them and their heterogeneous $SIQ$ networks are sparse. 
To cope with this issue, Learnlex will ask each newly-registered student to finish 15 starting questions that are carefully selected by question designers by considering the question popularity (i.e., how many prior students have worked on it), the grade of the student, and the question difficulty level. 
By combining the newly-registered student's interaction data on the starting questions and prior students' existing problem-solving interaction data, we build the initial heterogeneous $SIQ$ network for a new student and further train a R$^2$GCN model to predict his/her performances on other questions, which can achieve the same level of prediction accuracy as shown in Table~\ref{result_baseline2}.
% gain a good prediction accuracy 
% for a student at his/her early stage when working on Learnlex.
}

\haotian{
% Learnlex is one of their core products which has offered interactive math questions to more than 100,000 K-12 students from around 400 schools in both Hong Kong and Mainland China. 
% After the accomplishment of deployment in September, our method 
% % will be integrated into their online learning platform, which 
% will be integrated into their question pool to facilitate personalized learning path planning and question recommendation for students.
Once the testing and deployment is finally finished, our approach will work as the core component of question recommendation module on Learnlex, which can recommend appropriate questions that satisfy the requirements of different students.
% recommend appropriate questions
% % that better satisfy students' requirements 
% to different students on Learnlex.
% their personalized learning path planning engine which recommend and design different question-solving sequences for different students.
According to the market share estimation by Trumptech, Learnlex will be used by more than 100,000 students in the coming three years, which means that
over 100,000 students will benefit from the personalized online learning from our approach in the near future. 
% One senior engineer commented that \emph{``the prediction accuracy is impressive compared with other state-of-the-art approaches. It provides student performance prediction on each question, which can hardly be finished by prior knowledge tracing-based algorithms.''}
}

\vspace{-0.5em}
\section{Discussion}\label{discussion}
% This section will mainly discuss the generalizability of our method and the cold start problem.
% \vspace{-5pt}
\paragraph{Generalizability of our method}
Our method has the potential to be applied on touch screen devices since basic gestures
% \footnote{\href{https://developer.apple.com/design/human-interface-guidelines/ios/user-interaction/gestures/}{https://developer.apple.com/design/human-interface-guidelines/ios/user-interaction/gestures/}} and Android \footnote{\href{https://material.io/design/interaction/gestures.html}{https://material.io/design/interaction/gestures.html}} 
are similar to mouse movement interactions (e.g., Tap, Drag). However, there are some unique gestures on touch screens (e.g., Swipe, Flick), so we may need to add other features related to such gestures to extract more information. 
% from gesture sequences. 
\haotian{Also, our method can be applied to other online scenarios with rich mouse interactions of users (e.g., E-sports).} 
% For example, our method can be easily extended to analyze user behaviors and predict the performance of users in E-sports.
\vspace{-5pt}

% \paragraph{Explainability of GNNs}
% The main drawback of applying GNNs on student performance prediction is the limitation of GNNs's explainability since GNNs learn both topological information and node features. Researchers are still exploring the method of opening the "black boxes" of GNNs~\cite{gnnexplainer, gnnexplainer3}. However, those existing works can't explain our R$^2$GCN on a heterogeneous network. In Section \ref{experiment:topological}, we manage to interpret the results from the perspective of topology. To better provide some insights for educators, further explaining current prediction results from the aspect of node features is helpful.
% \vspace{-5pt}

\paragraph{Cold start problem}
\haotian{Cold start is a general problem for prediction and
a set of training data is needed to train the prediction model.}
% which means that the model cannot be used until there is sufficient data. 
Our method based on GNN also suffers from this problem. It requires that the questions and students have enough records.
% \haotian{To handle this issue, we ask the newly-registered students to finish 15 carefully-selected starting questions before R$^2$GCN is used to predict their performances on other questions, as discussed in Section~\ref{deployment}.
% Some more advanced methods can be further explored to handle this issue.}
% If this requirement is not fulfilled, one possible solution is to use extra information (e.g., the school of the student, question labels) to find similar questions or students to estimate the performance.
\haotian{To handle this issue, we ask the newly-registered students to finish 15 carefully-selected starting questions before R$^2$GCN is used to predict their performances on other questions, as discussed in Section~\ref{deployment}. Some more advanced method can be further explored to cope with it.}
% \vspace{-1em}
\section{Conclusion}

Student performance prediction is important for online education. However, most existing work targets at
% student performance prediction in 
MOOC platforms in which a predefined order of the learning materials (e.g., course videos and questions) is available. Little research has been done on interactive online question pools with no predefined question order.
In this paper, we propose a novel approach based on graph neural networks to predict students' score level on each question in interactive online question pools. Specifically, we formalize the student performance prediction in interactive online question pools as a node classification problem on a large heterogeneous network consisting of questions, students, and the interactions between them, better capturing the underlying relationship among questions and students. Then, we propose a novel GNN model R$^2$GCN by adapting the classical R-GCN model and further incorporating a residual connection between different convolutional layers. Our detailed evaluations on a real-world dataset collected from a K-12 mathematics online question pool show that our approach outperforms both traditional machine learning models (e.g., Logistic Regression (LR) and Gradient Boosted Decision Tree (GBDT) model) and the classical R-GCN model~\cite{Schlichtkrull2017ModelingRD}, in terms of accuracy and F1 score. 
% Despite the performance, our method has great generalizability since it can be easily applied on touch screen devices with minor augment on some unique gestures on touch screens, e.g., swipe and flick.

\haotian{In future work, we plan to extend the proposed student performance prediction approach using GNN to other 
% online learning platforms, for example, 
interactive online question pools designed for touch screen devices like iPad, and further evaluate our model on larger datasets collected with more students and questions.
Also, 
% besides student performance prediction, 
it will be interesting to further explore whether the proposed approach can be adapted to other analysis tasks in online education, e.g., detecting cheating behaviors.}
% of students in online learning.
% Furthermore, we would like to
% % improve the interpretability of the proposed GNN model, R$^2$GCN and 
% further include some methods to tackle the cold start problem. 
%
% Furthermore, we would like to further introduce new methods to better tackle the cold start problem.

% In future work, we would like to further utilize the framework proposed by us to explore other tasks including student clustering and cheating detection. 
% Also, one possible future work could be further explaining the GNN and features proposed in our work.
% \begin{acks}
% \end{acks}
\vspace{-1em}
\begin{acks}
This work is partially sponsored by Innovation and Technology Fund (ITF) with No. ITS/388/17FP. Y. Wang is the corresponding author. We would like to thank Trumptech (Hong Kong) Limited for providing the dataset and giving us feedback on the proposed method, Zhihua Jin for suggestions, Meng Xia and Min Xu for proofreading.
%  evaluation of our method and suggestions for future improvement. 
\end{acks}
% \vspace{-0.5em}
% \vspace{-1em}
\bibliographystyle{ACM-Reference-Format}
\bibliography{ref_new}

% \appendix
% \input{src/feature_list.tex}
% \input{src/evaluation.tex}









\end{document}